\documentclass{article}

    \PassOptionsToPackage{numbers, compress}{natbib}

 \usepackage[main, final]{neurips_2026}

\usepackage[utf8]{inputenc} 
\usepackage[T1]{fontenc}    
\usepackage{hyperref}       
\usepackage{url}            
\usepackage{booktabs}       
\usepackage{amsfonts}       
\usepackage{nicefrac}       
\usepackage{microtype}      
\usepackage{xcolor}         
\usepackage{graphicx}
\usepackage{subcaption}
\usepackage{xcolor} 
\usepackage{amsmath} 
\usepackage{amsthm}
\usepackage{multirow}
\usepackage{pifont}
\usepackage{wrapfig}
\definecolor{darkred}{rgb}{0.8, 0.0, 0.0}
\definecolor{darkgreen}{rgb}{0.0, 0.5, 0.0} 
\usepackage[ruled]{algorithm2e}
\newtheorem{proposition}{Proposition}

\newtheorem{lemma}{Lemma}
\newtheorem{remark}{Remark}
\newtheorem{assumption}{Assumption}

\ifodd 0
\newcommand{\rev}[1]{{\color{blue}#1}} 
\else
\newcommand{\rev}[1]{#1}
\fi

\title{Mobility Helps Learning: Unsupervised Model Adaptation for Object Recognition via Movement}

\author{%
    Yanan Ma, \qquad Yihang Tao, \qquad Zhengru Fang, \qquad  Senkang Hu\\
    Department of Computer Science, 
    ~City University of Hong Kong\\
    \texttt{\{yananma8-c, yihang.tommy, zhefang4-c, senkang.forest\}@my.cityu.edu.hk} \\    
    \And
    Xianhao Chen \\
    Department of Electrical and Computer Engineering \\
    The University of Hong Kong \\
    \texttt{xchen@eee.hku.hk} 
    \And
    Yuguang Fang\thanks{Corresponding author.} \\
    Department of Computer Science\\
    City University of Hong Kong\\
    \texttt{my.fang@cityu.edu.hk}
}

\begin{document}

\maketitle

\begin{abstract}

    Movable agents, such as autonomous vehicles and robots, should be able to autonomously adapt their pre-trained object recognition models to new environments without human annotations. While existing test-time adaptation (TTA) methods address label-free model adaptation, they typically treat test data as independent snapshots, without constructively exploiting a rich, unique source of supervision: the significant variance in prediction quality as an agent observes the same object from different distances and viewpoints.
    To capitalize on this, we propose \texttt{MoCaFe}, a model-agnostic and hyperparameter-insensitive paradigm that leverages motion-induced prediction variance for unsupervised adaptation. Specifically, we first develop mobility-calibrated filtering of pseudo-labels, which constructs reliable pseudo-labels by fusing cross-view predictions using inverse-variance weights. The filtered pseudo-labels then guide the selection of ``hard samples'' for effective learning. To further suppress pseudo-label corruption, we introduce \texttt{p-MoCaFe}, an optional extension that uses a public dataset to assign weighting factors to data samples, yielding an estimator of the clean loss with bounded bias. Extensive experiments on autonomous driving (nuScenes, KITTI) and embodied-agent datasets show that \texttt{MoCaFe} significantly outperforms state-of-the-art TTA schemes, achieving up to approximately 20\% accuracy gain.

\end{abstract}

\section{Introduction}
\label{sec:intro}
Object recognition models are notoriously weak at generalization, often suffering significant performance degradation even with minor changes in data distributions~\citep{recht2018cifar, koh2021wilds}. This limitation is particularly critical for mobile agents such as autonomous vehicles and robots, as their very purpose is to navigate and operate in a world that is, by nature, new and unpredictable. To bridge the generalization gap, recognition models often need to continually adapt to new environments~\cite{Samuel59,zhu2023pockengine}. However, traditional supervised adaptation is impractical in this context, since it relies on a labor-intensive and time-consuming manual labeling process~\cite{ma2025sense4fl}. 
For instance, adapting models locally on a vehicle would require drivers' manual labeling, which is often impractical. 
This creates a critical need for an automated, unsupervised adaptation framework with minimal human intervention.

\begin{figure}[t]
\centering
\begin{subfigure}{0.42\columnwidth}
    \centering
    \includegraphics[width=\linewidth]{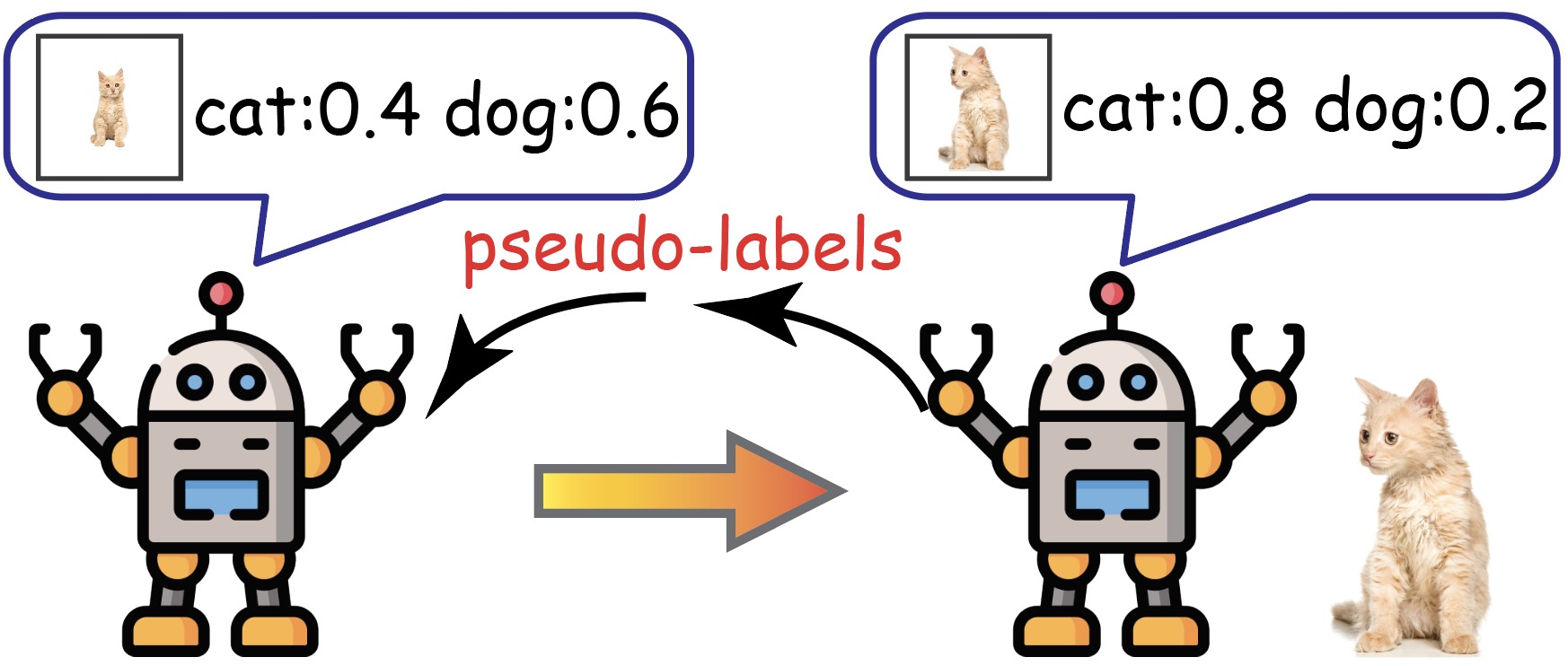}
    \caption{Illustration of mobility helps learning.}
    \label{fig:system_a}
\end{subfigure}
\hfill
\begin{subfigure}{0.48\columnwidth}
    \centering
    \includegraphics[width=\linewidth]{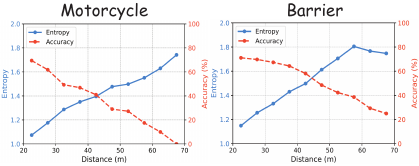}
    \caption{Accuracy and entropy vs. distance.}
    \label{fig:system_b}
\end{subfigure}
\caption{(a) The agent is more likely to produce the correct class, i.e., cat, when moving closer to the object. 
Intuitively, the high-confidence predictions (on the right) can serve as pseudo-labels to guide model adaptation over samples with lower confidence (on the left). 
(b) Average entropy and classification accuracy versus the distance for two classes of objects, i.e., barriers and motorcycles, in the nuScenes dataset with a pre-trained ResNet-50. 
Accuracy \textbf{varies significantly as agents move}.
}
\label{fig:system}
\end{figure}

To address this need, unsupervised model adaptation, or test-time adaptation (TTA), has emerged as a promising field~\cite{chen2022contrastive, sun2020test,kundu2020universal, liang2020we}. In the vision domain, existing TTA methods have shown promise, often by employing uncertainty-guided pseudo-labeling~\cite{Litrico_2023_CVPR}, image/feature generation~\cite{zhang2022memo}, and contrastive learning~\cite{chen2022contrastive, sun2020test}. However, the effectiveness of prior TTA methods is still hindered by two shortcomings. \ding{182} \textit{First}, existing approaches typically treat test data as independent snapshots, without constructively exploiting a rich and freely available source of information: a movable agent observes the same object instance spatiotemporally from varying distances and viewpoints. This motion naturally induces significant variance in prediction quality; an object that is ambiguous from afar is often easily classified up close, and a partially occluded object may become fully visible in subsequent frames  (Fig.~\ref{fig:system}). Consequently, high-quality predictions upon movement can naturally serve as ``teacher'' outputs\footnote{In this paper, we refer to the constructed pseudo-label used for supervising model training as teacher outputs, and the training data samples guided by these outputs as student samples.} without manual labeling. Note that, although leveraging mobility to learn has been exploited in the semi-/unsupervised learning context~\cite {agrawal2015learning, zhou2017unsupervised}, these schemes are mostly focused on representation learning rather than unsupervised model adaptation, particularly for object recognition. \ding{183} \textit{Second}, prior methods are often hyperparameter-sensitive or model-specific~\cite{yuan2023robust}. A framework meticulously tuned for one environment or architecture may fail or require careful re-tuning when applied to another, rendering it unreliable for real-world deployment. Consequently, two research questions remain underexplored in TTA for object recognition tasks:

\begin{itemize}
    \item \textit{\textbf{Q1:} How can we effectively harness movement as a source of supervision for object recognition without manual annotations?}
    \item \textit{\textbf{Q2:} How can we design a model-agnostic, hyperparameter-insensitive, and robust adaptation framework that exploits mobility-induced information?}
\end{itemize}

To answer the research questions above, this paper proposes \texttt{MoCaFe}, a novel TTA paradigm for object recognition that leverages motion-induced prediction variance for model adaptation. We treat cross-view predictions as noisy observations of an underlying ground-truth state and fuse them into a mobility-calibrated pseudo-label. A data-sampling process is then employed to select student samples for learning with large cross-entropy relative to the filtered pseudo-label, facilitating effective learning via movement (answering \textbf{Q1}). Importantly, the filtered pseudo-label relies on a simple, inverse-variance prediction combination, which is hyperparameter-insensitive, model-agnostic, and robust to model over-confidence (answering \textbf{Q2})\footnote{Framework-induced hyperparameters exclude common hyperparameters, such as learning rates, which are needed by general learning frameworks.}. We further develop \texttt{p-MoCaFe}, an optional calibrated extension based on a public dataset, and theoretically show that its calibrated objective yields a bounded-bias estimator of the clean loss. Our contributions are summarized as follows.

\begin{itemize}
    \item We present the idea of ``mobility helps learning'', a mobility-driven TTA paradigm that exploits the fact that movable agents naturally observe the same object from multiple distances and viewpoints. 
    This enables cross-view self-supervision without requiring manual labels from the target adaptation environment.

    \item We propose \texttt{MoCaFe}, a mobility-calibrated filtering scheme for pseudo-labeling inspired by information filtering. This framework is coupled with a cross-entropy-based data sampling process to selectively learn from informative student samples.

    \item We develop \texttt{p-MoCaFe}, an optional calibrated extension for \texttt{MoCaFe} based on a public dataset. We theoretically show that the calibrated objective yields a bounded-bias estimator of the clean loss.

    \item We conduct extensive experiments across different model architectures and datasets. 
    The results show that both \texttt{MoCaFe} and \texttt{p-MoCaFe} significantly improve adaptation performance, with up to approximately 20\% accuracy gain for transformer-based models over strong TTA baselines, while maintaining robustness across hyperparameter settings and validation dataset choices.
\end{itemize}

\section{Related Work}

\textbf{Test-Time Adaptation.}
TTA adapts models to unlabeled data without requiring access to the source dataset~\cite{chen2022contrastive,Litrico_2023_CVPR,yuan2023robust,karim2023c,chen2024uncertainty,joo2024improving, ma2026learning}. In \cite{sun2020test}, test-time training introduces a self-supervised auxiliary rotation prediction task to be optimized jointly during both source and target training. However, this approach requires modification to pre-training. In \cite{chen2022contrastive}, Chen \textit{et al.} propose to refine predictions within a self-supervised strategy. 
In \cite{Litrico_2023_CVPR}, Litrico \textit{et al.} propose a framework that reweights classification loss according to pseudo-label uncertainty to improve robustness by progressively refining the pseudo labels via neighborhood knowledge aggregation. Nevertheless, the aforementioned works treat test data as independent samples. In \cite{gong2022note}, by observing that data can be temporally correlated in application scenarios, e.g., autonomous driving, Gong \textit{et al.} propose a scheme against such non-IID data streams with temporal correlation. However, while the time correlation of data frames upon agent movement is considered, their goal is to combat its negative effects rather than exploit it as a constructive source for adaptation. Overall, prior TTA methods mainly rely on model-internal supervisory signals on \textit{individual} samples (i.e., pseudo-labeling or entropy minimization) without \textit{constructively} leveraging the cross-view prediction relationships and variances. As a result of their design, most existing TTA approaches remain sensitive to hyperparameter choices or model architecture~\cite{boudiaf2022parameter}. Our framework stands out by not only exploiting a novel dimension of design, i.e., harnessing movement as a source of supervision for object recognition, but also delivering superior performance while being model-agnostic and hyperparameter-insensitive.

\textbf{Other Unsupervised Learning Approaches.} 
Apart from TTA, there are some other unsupervised learning frameworks, yet they are designed for different goals. \textbf{Contrastive learning} is a powerful unsupervised learning paradigm that aims to learn an embedding space~\cite{chuang2020debiased}. A seminal work in this field is InfoNCE \cite{oord2019representationlearningcontrastivepredictive}, which introduces a contrastive loss function to compare positive samples against multiple negatives. Many recent popular frameworks, such as DINO and iBOT, exploit multiple views of the original images (e.g., by view crops or masking) via \textit{data augmentation} for contrastive learning~\cite{caron2021emerging,zhouimage,oquab2024dinov2,bachman2019learningrepresentationsmaximizingmutual,he2019moco}. Conceptually similar, Agrawal \textit{et al.} propose a feature learning approach based on egomotion of agents~\cite{agrawal2015learning}. Nevertheless, these contrastive learning approaches aim to learn transferable features during the \textit{pre-training stage}, which are then applied to specific downstream tasks using labeled data. This fundamentally differs from our approach, which targets the model adaptation (\textit{fine-tuning}) stage for object recognition, and thus, faces unique challenges in pseudo-label generation and label-noise mitigation.

\rev{\textbf{Knowledge distillation} (KD)~\cite{hinton2015distilling} is increasingly applied to unlabeled datasets by employing a powerful teacher model to generate pseudo labels~\cite{Park_2019_CVPR}.} In \cite{iliopoulos2022weighted}, a ``debiasing'' reweighting framework is proposed to reweight the loss function of a student model on an unlabeled dataset. In \cite{kontonis2024slam}, a student-label mixing (SLaM) approach is proposed to modify the student's loss by incorporating the teacher's noise model. However, our approach differs from these label-free KD mechanisms because both ``teacher'' and ``student'' predictions \textit{come from the same pre-trained model}.


\section{Problem Definition}
\label{sec: classification}
We consider a classical multiclass classification problem.
Let $\mathcal{X}$ denote the input space and $\mathcal{Y} = \{1, \ldots, C\}$ represent the label space with $C$ distinct classes. Our training dataset $\mathcal{S} = \{(x_1, y_1), (x_2, y_2), \ldots, (x_n, y_n)\}$ consists of $n$ samples drawn from an unknown distribution $\Pi$ over $\mathcal{X} \times \mathcal{Y}$. 
The task is to learn a logit predictor
$f:\mathcal{X}\to\mathbb{R}^C$, whose predicted class is
$\arg\max_c [f(x)]_c$, that minimizes the risk associated
with misclassification. To measure the performance of $f$, we define the expected risk
\begin{equation}
   R(f) = \mathbb{E}[\ell(y, f(x))],\label{loss}
\end{equation}
where $(x, y)$ is sampled from $\Pi$ and $\ell(\cdot)$ is a loss function.
We aim to find the function $f^\star$ that minimizes the empirical risk
\begin{equation}
    f^\star = \arg\min_{f \in \mathcal{F}} \frac{1}{n} \sum_{(x, y) \in S} \ell(y, f(x)),
\end{equation}    
where $\mathcal{F}$ is a predefined hypothesis class of predictors, often parameterized by their weights. 

However, as can be seen, the above classical optimization demands ground-truth label $y$, which is assumed to be unavailable in our case. In what follows, we propose the \texttt{MoCaFe}  framework to effectively optimize $f$ by exploiting the different spatial-temporal viewpoints on object instances.


\section{Proposed \texttt{MoCaFe} Framework}
\label{sec: framework}

In this section, we first elaborate on our fully unsupervised \texttt{MoCaFe} framework (Section \ref{subsec: mocafe}), which constructs filtered pseudo-labels by aggregating multiple predictions of the same object according to their estimated observation reliability. We then introduce \texttt{p-MoCaFe} (Section \ref{subsec:p-MoCaFe}), an optional extension that utilizes a public dataset to achieve a bounded-bias estimator of the clean loss.

\subsection{\texttt{MoCaFe} Scheme}\label{subsec: mocafe}

\begin{figure*}[t]
\centering
\includegraphics[width=\textwidth]{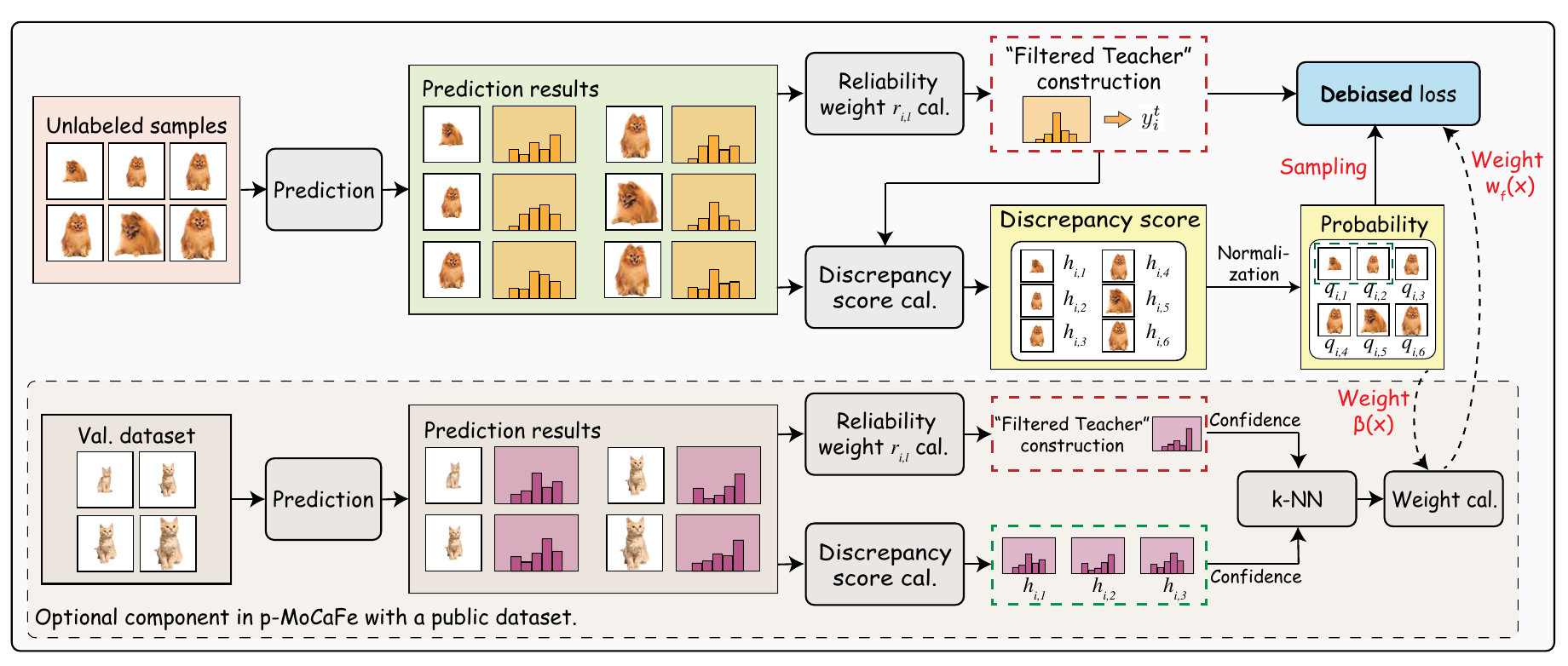}
\caption{
\rev{\textbf{Overview of the \texttt{MoCaFe} and \texttt{p-MoCaFe} frameworks detailed in Section \ref{sec: framework}.} Multi-view predictions are aggregated via reliability weighting to construct a robust "Filtered Teacher" pseudo-label. Discrepancy scores against this teacher are then used to select the most informative student samples for training. In the p-MoCaFe extension, an additional public dataset is used to calibrate debiasing weights, yielding an unbiased empirical risk for robust model adaptation.
}
}
\label{fig:mocafe}
\end{figure*}

\subsubsection{Reliability-Weighted Pseudo-Label Construction}

Let $\mathcal{L}_{i}=\{x_{i,1}, \ldots, x_{i,L_{i}}\}$ denote a sequence of $L_i \geq 2$ frames containing the same object $i$, which can be obtained based on temporal correlation and class-agnostic object-tracking methods. We aim to construct a pseudo-label from each sequence $\mathcal{L}_{i}$. To theoretically justify our framework, we make the following assumption~\cite{kendall2017uncertainties}.


\begin{assumption}\label{assumption: logit model}
    We assume that the multi-view logit predictions for an object $i$ can be modeled as noisy observations of a latent semantic state. Specifically, the logit vector $\mathbf{z}_{i,l} \in \mathbb{R}^{C}$ for frame $l$ is formulated as
    \begin{equation}
        \mathbf{z}_{i,l} = \mathbf{s}_{i} + \boldsymbol{\epsilon}_{i,l}, \quad \boldsymbol{\epsilon}_{i,l} \sim \mathcal{N}(\mathbf{b}_{i,l}, \sigma^{2}(u_{i,l})\mathbf{I}),
        \label{eq:mcf_observation_model}
    \end{equation}
    where $\mathbf{s}_{i}$ is the true state and $\boldsymbol{\epsilon}_{i,l}$ is stochastic noise. The mean of the noise $\mathbf{b}_{i,l}$ captures prediction bias (e.g., from domain shift or occlusion), while the variance $\sigma^2(u_{i,l})$ is conditioned on an uncertainty score $u_{i,l}$\footnote{\rev{Here, $u_{i,l}$ is set to the normalized predictive entropy: $u_{i,l} = -\frac{1}{\log C} \sum_{c=1}^C p_{i,l,c} \log p_{i,l,c}$ without requiring ground-truth labels. If distance measurements $d_{i,l}$ are available (e.g., via radar), the variance can be extended to $\sigma^2(d_{i,l}, u_{i,l})$.}}.
\end{assumption}


\rev{To estimate observation variance, we partition the uncertainty space into discrete bins defined by $\mathcal{G}_k = \{(i,l): u_{i,l} \in U_k\}$, where $U_{k}$ denotes the uncertainty interval}\footnote{\rev{With available distance measurements, this can be extended to a 2D mobility-uncertainty partition: $\mathcal{G}_{k}=\{(i,l) : d_{i,l} \in D_{j}, u_{i,l} \in U_{k}\}$, where $D_{j}$ is the distance interval. Since distance is solely used for interval assignment, coarse range estimates (e.g., 0–40 m, 40–80 m, >80 m) are sufficient.}}.
The variance for $\mathcal{G}_{k}$ is empirically estimated as 
\begin{equation}\label{eq:residual_inconsistency}
\hat{\sigma}_{k}^{2} = \frac{1}{|\mathcal{G}_{k}|}\sum_{(i,l)\in\mathcal{G}_{k}}\rho_{i,l}, ~~~\mathrm{with} ~~
    \rho_{i,l} = \frac{1}{C}\left\|\mathbf{z}_{i,l} - \widehat{\mathbf{s}}_{i,-l}\right\|_2^2,
\end{equation}
where $\widehat{\mathbf{s}}_{i,-l} = \frac{1}{L_i-1} \sum_{\tau\neq l}\mathbf{z}_{i,\tau}$. Here, $\rho_{i,l}$ is the residual inconsistency of frame $l$ based on \textit{leave-one-out} sequence consensus. By mapping this back to individual observations, $\widehat{\sigma}^2(u_{i,l}) = \widehat{\sigma}_k^2$, we define the reliability weight for each frame via inverse-variance weighting
\begin{equation} \label{eq:reliability}
   r_{i,l} = \frac{1}{\hat{\sigma}^{2}(u_{i,l}) + \varepsilon_{r}}, 
\end{equation}
where $\varepsilon_{r}>0$ is a small stabilizing constant.
\rev{Then, we aggregate the weighted logits to construct the filtered pseudo-label vector $\boldsymbol{\mu}_{i}$, yielding the sequence-level pseudo-label $y_{i}^{t}$}
\begin{equation}\label{eq:filtered_teacher}
    \boldsymbol{\mu}_{i} = \frac{\sum_{l=1}^{L_{i}} r_{i,l}\mathbf{z}_{i,l}}{\sum_{l=1}^{L_{i}} r_{i,l}}, \quad y_{i}^{t} = \arg\max_{c \in \{1,\dots,C\}} [\boldsymbol{\mu}_{i}]_c.
\end{equation}
Unlike standard confidence-based selection, this formulation relies on a reliability-weighted aggregation of multiple observations. Predictions with lower residual variance are assigned higher weights and contribute more to the filtered pseudo-labels.

\textbf{Theoretical Analysis.}
We establish the following theoretical results to justify our framework. For notational simplicity, we drop the object-specific index $i$ and consider a generic sequence length $L$.

\begin{lemma}\label{lemma: variance reduction}
    The stochastic posterior variance $\Sigma_L = \left( \sum_{l=1}^L \sigma_l^{-2} \right)^{-1}$ of the filtered prediction is lower than that of any individual observation:
    \begin{equation}
        \Sigma_L < \min_{1 \le l \le L} \sigma_l^2. 
    \end{equation}
\end{lemma}

\begin{proposition}\label{proposition: error_bound}
    Assume $\mathbf{b}_l = \mathbf{0}$ for all $l$, with Lemma \ref{lemma: variance reduction}, the theoretical misclassification upper bound of the filtered pseudo-label $y^t$ is lower than that of the single best prediction $y_{best}$:
    \begin{equation}
        \mathbb{P}(y^t \neq c^\star) \le \sum_{c \neq c^\star} \exp\left(-\frac{\Delta_{0,c}^2}{4\Sigma_L}\right) < \sum_{c \neq c^\star} \exp\left(-\frac{\Delta_{0,c}^2}{4\sigma_{best}^2}\right),
    \end{equation}
    where $c^\star$ is the ground-truth class, and $y_{best}$ denotes the prediction from the single best observation corresponding to the minimum variance $\sigma_{best}^2 = \min_{1 \le l \le L} \sigma_l^2$.
\end{proposition}

The detailed proofs for Lemma \ref{lemma: variance reduction} and Proposition \ref{proposition: error_bound} are provided in Appendix \ref{app: variance reduction} and \ref{app:error_bound}.

\begin{remark}
    Proposition \ref{proposition: error_bound} demonstrates that the upper bound of the filtered teacher is smaller than the upper bound of any single ``best'' prediction, assuming bias $\mathbf{b}_l = \mathbf{0}$ (i.e., the prediction is the ground-truth prediction with zero-mean Gaussian noise, as in ~\cite{kendall2017uncertainties}). While this assumption may not always hold, it provides theoretical justification for the design of the filtered pseudo-labels. Empirically, our experiments show that the filtered pseudo-label is more robust than simply selecting the highest-confidence prediction from each sequence as the teacher (see Appendix \ref{app:ablation}). 
\end{remark}

\subsubsection{Informative Sampling with Debiasing}
\label{subsubsec: data_selection}

To facilitate efficient adaptation, we preferentially sample informative frames that ``disagree'' with the filtered pseudo-label, which are essentially ``hard examples''. Specifically, we define a discrepancy score between the filtered pseudo-label and the individual prediction
$h_{i,l} = \ell(y_i^t, f(x_{i,l}))$,
where $\ell$ denotes the loss (e.g., cross-entropy). A larger $h_{i,l}$ indicates high inconsistency, suggesting that the frame should be selected for adaptation. We then construct sampling probabilities
\begin{equation}
    q_{i,l} = \frac{h_{i,l}+\eta}{\sum_{i'=1}^{N}\sum_{\tau=1}^{L_{i'}}\left(h_{i',\tau}+\eta\right)}, ~~ \forall i, \forall l, \label{eq:mcf_sampling_probability}
\end{equation}
where $\eta>0$ is a small value preventing zero-probability sampling for frames already aligned with pseudo-labels. Accordingly, we select $m$ samples from the dataset, forming the selected dataset $\mathcal{X}_{m}$.

Since the sampling process above draws from an instrumental distribution $\mathbb{Q}$, it may not perfectly match the true data distribution $\Pi$, potentially introducing bias. Assuming the filtered pseudo-label is perfect, this empirical risk can be easily debiased by weighting factors $\beta(x_{i,l})$ via
\begin{equation}\label{eq:weighted_sampling}
    \hat{R}_q(f) = \frac{1}{m} \sum_{x_{i,l} \in \mathcal{X}_m} \beta(x_{i,l}) \ell(y_i^t, f(x_{i,l})).
\end{equation}

Assuming $y^t(x)$ is the ground-truth label, i.e., $y^t(x) = y$, we have the following proposition.
\begin{proposition} \label{prop:sampling}
    If $\beta(x)={\pi\left(x\right)}/{q\left(x\right)}$, where $\pi\left(x\right)$ is the probability density of the distribution $\Pi$ and $q\left(x\right)$ is the probability density of the instrumental distribution $\mathbb{Q}$, then $\hat{R}_{q}(f)$ in \eqref{eq:weighted_sampling} is an unbiased estimator of $R(f)$.
\end{proposition}

Note that $\beta(x) \propto {1}/{q\left(x\right)}$ if $x$ is drawn i.i.d. from $\Pi$.
Proposition \ref{prop:sampling} is a direct consequence of the definition of $\beta(x)$, as shown in \cite{kanamori2003active}. 
The overall \texttt{MoCaFe} framework is illustrated in Fig.~\ref{fig:mocafe} and summarized in Alg.~\ref{alg:mocafe_framework}.

   
   
   

\begin{algorithm}[t]
   \caption{The \texttt{MoCaFe} Framework}
   \label{alg:mocafe_framework}
   \LinesNumbered
   \KwIn{Training dataset $\mathcal{X}$ composed of tracked sequences $\{\mathcal{L}_i\}$, pre-trained model $f$}
   \KwOut{Adapted model $f^\star$}

\For{each sequence $\mathcal{L}_i \in \mathcal{X}$}{
    \For{each frame $x_{i,l} \in \mathcal{L}_i$}{
        Compute logits $\mathbf{z}_{i,l} = f(x_{i,l})$ and uncertainty
        $u_{i,l}$\;
    }
    Compute 
    $\rho_{i,l}$
    for all frames in $\mathcal{L}_i$
    via Eq.~\eqref{eq:residual_inconsistency}\;
}

   Partition all observations $\{(i,l)\}$ into uncertainty bins $\{\mathcal{G}_k\}_{k=1}^K$ based on $u_{i,l} \in U_k$\;
   \For{each bin $\mathcal{G}_k$}{
       Compute pooled empirical variance $\widehat{\sigma}_k^2 = \frac{1}{|\mathcal{G}_k|} \sum_{(i,l) \in \mathcal{G}_k} \rho_{i,l}$ via Eq.~\eqref{eq:residual_inconsistency}\;
       Compute bin reliability weight $r_k = \frac{1}{\widehat{\sigma}_k^2 + \varepsilon_r}$ via Eq.~\eqref{eq:reliability}\;
   }
   Assign frame reliability $r_{i,l} \leftarrow r_k$ for each $(i,l) \in \mathcal{G}_k$\;
   
   \For{each sequence $\mathcal{L}_i \in \mathcal{X}$}{
       Construct $\boldsymbol{\mu}_i$ and $y_i^t$ via Eq.~\eqref{eq:filtered_teacher} and 
assign $y^t(x_{i,l}) \leftarrow y_i^t$ and $\boldsymbol{\mu}(x_{i,l}) \leftarrow \boldsymbol{\mu}_i$\;
   }
   $\mathcal{Y}^t \leftarrow \{y^t(x) \mid x \in \mathcal{X}\}$\;
   Calculate 
    $h(x) = \ell(y^t(x), f(x))$ for $x \in \mathcal{X}$\;
    Compute 
$q(x) \leftarrow
\frac{h(x) + \eta}
{\sum_{x' \in \mathcal{X}} (h(x') + \eta)}$
for all $x \in \mathcal{X}$; Sample $\mathcal{X}_m \subset \mathcal{X}$ according to $q(x)$\;
   Adapt model $f$ over $\mathcal{X}_m$ by minimizing $\hat{R}_q(f)$ via Eq.~\eqref{eq:weighted_sampling}\;
\end{algorithm}


\subsection{\texttt{p-MoCaFe}: An Optional Extension Using Public Datasets}
\label{subsec:p-MoCaFe}

The \texttt{MoCaFe} framework yields an unbiased estimator of $R(f)$ only if the filtered pseudo-labels $y^t(x)$ constitute the ground truth. Since it inevitably contains inaccuracies, we introduce \texttt{p-MoCaFe}, an optional extension using a public dataset to debias the empirical risk via data weighting.
In particular, after the whole process of \texttt{MoCaFe}, \texttt{p-MoCaFe} replaces weight $\beta(x)$ in (\ref{eq:weighted_sampling}) with $w_f(x)$
\begin{equation}
    \begin{aligned} \label{eq:weights}
        w_f(x)=\frac{\beta(x)}{1+p(x)\left(d_f(x)-1\right)}, \quad \mathrm{with} ~~
    d_f(x)=\frac{\ell\left(f_{\mathrm{false}}(x), f(x)\right)}{\ell\left(y, f(x)\right)}.
    \end{aligned}
\end{equation}
The weighted empirical risk, therefore, can be expressed as 
\begin{equation} \label{eq:weighted_risk}
    R^w(f)=\frac{1}{m} \sum_{x \in \mathcal X_m} w_f(x) \ell(y^t(x), f(x)).
\end{equation}
From Eq. \eqref{eq:weights}, we can observe that the weight for each sample $x$ depends on two factors: \textit{(i) the likelihood that the teacher corrupts its label, and (ii) the extent to which this corrupted label distorts the risk.} Specifically, if the probability of a corrupted label is high or the distortion term is large, we should assign a smaller weight to minimize the influence of the teacher's erroneous prediction \rev{(See Appendix \ref{proof:MSE})}.
We further show that the weighted empirical risk $R^w(f)$ in Eq. \eqref{eq:weighted_risk} is an unbiased estimation of $R(f)$.

\begin{proposition} \label{prop:weighted_risk}
    Assume that $\mathcal{X}_m$ is drawn from the induced distribution $\mathbb{\tilde Q}_{\text{t}}$, then we have:
   \begin{equation}
       \mathbb{E}\left[R^w(f)\right]=R(f).
   \end{equation}
\end{proposition}
The detailed proof follows from \cite{iliopoulos2022weighted}, which can be found in Appendix \ref{proof:weighted_risk}. 
Given a sufficiently large dataset sampled from $\mathbb{\tilde Q}_{\text{t}}$, optimizing an unbiased estimator for the risk should provide a better approximation for $\min_{f \in \mathcal{F}} R(f)$ than optimizing an estimator with constant bias.

\textbf{Estimating Weights (\ref{eq:weights}).}
To calculate the weights in Eq. (\ref{eq:weights}), one hurdle is that $p(x)$ and $d_f(x)$ are usually unknown. To address this issue, we assume the existence of a public dataset $\mathcal{S}_\text{v}=\left\{\left(x_1, y_1\right), \left(x_2, y_2\right), \ldots, \left(x_v, y_v\right)\right\}$. We apply the multi-view filtering process to the validation sequences to construct pseudo-labels $y^t(x)$, and the remaining frames are designated as student images.
We then estimate the weights $\hat{p}(x)$ and $\hat{d}_f(x)$ via the $k$-Nearest Neighbors ($k$-NN) method (see Appendix \ref{app:estimation} for details). Since the $k$-NN approach introduces approximation errors, in what follows, we generalize Proposition \ref{prop:weighted_risk} and the results in \cite{iliopoulos2022weighted}
by establishing that the deviation of our weighted empirical risk from the true risk remains bounded by these estimation errors.

\begin{proposition}\label{prop: bounded bias}
    The weighted empirical risk estimator, given by $\hat{R}^{\tilde{w}}(f) = \frac{1}{m}\sum_{x \in \mathcal{X}_m} \tilde{w}_f(x) \ell(y^t(x), f(x))$, satisfies:
    \begin{equation}
    \begin{aligned}
        \big|\mathbb{E}[\hat{R}^{\tilde{w}}(f)] - R(f)\big| &\le \mathbb{E}_{x \sim \mathbb{\tilde Q}_{\text{t}}}\Big[ \tilde{w}_f(x) \ell(y, f(x)) \big|p(x)(d_f(x)-1) - \hat{p}(x)(\hat{d}_f(x)-1)\big| \Big] \\
        &\quad + \mathbb{E}_{x \sim \mathbb{\tilde Q}_{\text{t}}}\Big[ \big|\tilde{w}_f(x) - \hat{w}_f(x)\big| \big[1+\hat{p}(x)(\hat{d}_f(x)-1)\big] \ell(y, f(x)) \Big],
    \end{aligned}         
    \end{equation}
    where $\hat{w}_f(x)$, $\hat{p}(x)$, and $\hat{d}_f(x)$ denote the raw estimates via $k$-NN, and $\tilde{w}_f(x) = \min\{\hat{w}_f(x), 1\}$ denotes the clipped weight.
\end{proposition}

The detailed proof is provided in Appendix \ref{app: bounded bias}. Proposition \ref{prop: bounded bias} reveals that the risk of \texttt{p-MoCaFe} framework contains two parts: estimation bias (the first term), which scales proportionally with the $k$-NN approximation error, and clipping bias (the second term), which arises when the raw estimated weight is truncated by a boundary, e.g., 1. Notably, we show in Appendix \ref{app:ablation} that \texttt{p-MoCaFe} consistently outperforms \texttt{MoCaFe} even with a substantial domain gap between the public and target datasets.
The \texttt{p-MoCaFe} framework is summarized in Alg. \ref{alg:p_mocafe} in Appendix \ref{app:algorithm}.


\section{Experiments} \label{sec:simulation}


\subsection{Experimental Settings}
\label{subsec:setup}

\textbf{Dataset.} We conduct experiments using autonomous driving datasets (nuScenes \cite{caesar2020nuscenes} and KITTI \cite{Geiger2013IJRR}), as well as embodied agent datasets (selected from JRDB \cite{martin2021jrdb}, MOT \cite{voigtlaender2019mots}, and LaSOT \cite{fan2019lasot}).
Since these datasets are captured by movable agents in the real world, they naturally provide images of the same object from varying distances/viewpoints and with different levels of confidence and accuracy. 
\rev{We construct instance sequences by extracting the same objects across images using ground-truth associations to demonstrate the theoretical and empirical potential of our method. For practical evaluation, we also report performance using an off-the-shelf class-agnostic object tracker.}

For all experiments, we pre-train these models in Singapore Onenorth on the nuScenes dataset. We test i) \textbf{cross-region} adaptation, which adapts the pre-trained model in the Boston Seaport scenario on the nuScenes dataset, ii) \textbf{cross-dataset} adaptation, which adapts the pre-trained model on the KITTI dataset, and iii) \textbf{cross-domain} adaptation, which adapts the pre-trained model on the embodied agent dataset. 
The latter two are provided in Appendix \ref{app:kitti} and \ref{app:Embodied}, which consistently demonstrate our schemes outperform other baselines.

\textbf{Implementation Details.} 
We adopt two widely used model architectures: a convolutional neural network (ResNet-50) and a transformer (Swin-T)~\cite{liu2022swin}. We also demonstrate that our method can be extended to adapt the classification part of Faster R-CNN, a well-known object detection model. 
\rev{We first perform inference over the unlabeled target sequences using the source-pretrained checkpoint to obtain the predictions (logits). Note that the varying pre-trained model accuracies (i.e., 75\%, 70\%, and 65\%) are evaluated directly on this target dataset. We then fine-tune the model over $\mathcal{X}_m$ using the weighted objective in Eq. \eqref{eq:weighted_sampling} for \texttt{MoCaFe} and Eq. \eqref{eq:weighted_risk} for \texttt{p-MoCaFe}. The adaptation process runs for 50 epochs for ResNet-50 and Swin-T, and 200 epochs for Faster R-CNN.} For ResNet-50, the learning rate is $10^{-5}$ for adaptation and $10^{-4}$ for pre-training, a momentum coefficient of 0.9, and a weight decay of $5\times10^{-4}$. 
The Swin-T configuration is adopted with a window size of 7 and a patch size of 4. The base learning rate is set to $10^{-5}$ with a 5-epoch warmup. 
We use the Singapore Onenorth scenario in the nuScenes dataset as the public dataset, with the size of $v=8000$.
\textit{Note that these hyperparameters are not induced by our framework}. For hyperparameter $k$ induced by \texttt{p-MoCaFe}, we set $k = \lceil {\sqrt{v}}/{2}\rceil$ and provide ablation analysis in Section \ref{subsec:ablation}.

\begin{table*}[t!]
\centering
\caption{Performance in the Boston Seaport scenario on the nuScenes dataset under varying pre-trained model accuracy (i.e., 75\%, 70\%, and 65\%). The value in the bracket represents the standard deviation. \textbf{Boldface} and \underline{Underline} highlight the highest and second-highest accuracy, except for the FSL benchmark. $\color{green}\uparrow$ and $\color{red}\downarrow$ indicate performance improvement/degradation.}
\label{table_pre_trained}
\resizebox{0.94\textwidth}{!}{
\renewcommand{\arraystretch}{1}
\begin{tabular}{l|ccc|ccc}
\toprule
& \multicolumn{3}{c}{\textbf{ResNet-50}} & \multicolumn{3}{|c}{\textbf{Swin-T}}\\
Pre-trained & 75\% & 70\% & 65\% & 75\% & 70\% & 65\%  \\
\midrule
w/o MoCaFe   & 76.18$\color{green}\uparrow$($\pm$  0.28) &72.05$\color{green}\uparrow$($\pm$  0.28)&62.17$\color{red}\downarrow$($\pm$  0.37) & 76.69$\color{green}\uparrow$($\pm$  0.26) & 72.55$\color{green}\uparrow$($\pm$  0.25) & 40.23$\color{red}\downarrow$($\pm$  0.43)    \\
AdaCon &  77.88$\color{green}\uparrow$($\pm$  0.19)  & 73.63$\color{green}\uparrow$($\pm$  0.15)  &  66.51$\color{green}\uparrow$($\pm$  0.18) & 76.85$\color{green}\uparrow$($\pm$  0.33)     & 72.83$\color{green}\uparrow$($\pm$  0.47)     & 62.28$\color{red}\downarrow$($\pm$ 0.61)     \\
PLUE &  77.94$\color{green}\uparrow$($\pm$  0.12)  & 73.82$\color{green}\uparrow$($\pm$  0.21)     & 67.28$\color{green}\uparrow$($\pm$  0.29)    &  77.43$\color{green}\uparrow$($\pm$  0.24)     & 73.55$\color{green}\uparrow$($\pm$  0.30)     & 59.91$\color{red}\downarrow$($\pm$ 0.35)     \\
RoTTA & 78.67$\color{green}\uparrow$($\pm$  0.21) & 75.13$\color{green}\uparrow$($\pm$  0.33) & 68.20$\color{green}\uparrow$($\pm$  0.31) & ----- & ----- & -----\\
C-SFDA & 78.71$\color{green}\uparrow$($\pm$  0.22) & 75.37$\color{green}\uparrow$($\pm$  0.26) & 68.34$\color{green}\uparrow$($\pm$  0.34) & 77.75$\color{green}\uparrow$($\pm$  0.18) & 72.20$\color{green}\uparrow$($\pm$  0.19) & 66.43$\color{green}\uparrow$($\pm$  0.25) \\
UPA & 76.87$\color{green}\uparrow$($\pm$  0.36) & 74.13$\color{green}\uparrow$($\pm$  0.34) & 57.60$\color{red}\downarrow$($\pm$  0.48) & 78.24$\color{green}\uparrow$($\pm$  0.26) & 73.15$\color{green}\uparrow$($\pm$  0.23) & 68.12$\color{green}\uparrow$($\pm$  0.28)\\
AnCon & 79.23$\color{green}\uparrow$($\pm$  0.29) & 75.61$\color{green}\uparrow$($\pm$  0.36) & 68.23$\color{green}\uparrow$($\pm$  0.38)& 
80.38$\color{green}\uparrow$($\pm$  0.27) & 
76.28$\color{green}\uparrow$($\pm$  0.30) & 70.59$\color{green}\uparrow$($\pm$  0.32)\\
\textbf{MoCaFe} & 
\underline{80.81$\color{green}\uparrow$($\pm$  0.23)} & \underline{77.25$\color{green}\uparrow$($\pm$  0.25)} & \underline{69.32$\color{green}\uparrow$($\pm$  0.24)} &  \underline{89.60$\color{green}\uparrow$($\pm$  0.11)} & \underline{89.34$\color{green}\uparrow$($\pm$  0.15)} & \underline{89.27$\color{green}\uparrow$($\pm$  0.16)} \\
\textbf{p-MoCaFe}  &  
\textbf{81.23$\color{green}\uparrow$($\pm$  0.13)}& \textbf{77.93$\color{green}\uparrow$($\pm$  0.16)}& \textbf{69.74$\color{green}\uparrow$($\pm$  0.25)}&  \textbf{90.19$\color{green}\uparrow$($\pm$  0.17)} & \textbf{90.07$\color{green}\uparrow$($\pm$  0.18)} & \textbf{90.02$\color{green}\uparrow$($\pm$  0.12)} \\
FSL & 88.66$\color{green}\uparrow$($\pm$  0.12) & 87.83$\color{green}\uparrow$($\pm$  0.18) & 86.87$\color{green}\uparrow$($\pm$  0.25)  & 92.30$\color{green}\uparrow$($\pm$  0.06) & 92.12$\color{green}\uparrow$($\pm$  0.15) & 91.98$\color{green}\uparrow$($\pm$  0.28)\\
\bottomrule
\end{tabular}
}
\end{table*}

\subsection{Comparison with Benchmarks}
\label{sec:benchmarks}

\textbf{Baselines.}
We compare our proposed \texttt{MoCaFe} and \texttt{p-MoCaFe} schemes with the following benchmarks: 
\textbf{AdaCon} \cite{chen2022contrastive},  
\textbf{PLUE} \cite{Litrico_2023_CVPR},
\textbf{RoTTA} \cite{yuan2023robust},
\textbf{C-SFDA} \cite{karim2023c},
\textbf{UPA} \cite{chen2024uncertainty}, and
\textbf{AnCon} \cite{joo2024improving}. 
We also include two benchmarks: \textbf{w/o MoCaFe}, a baseline employing uniform weighting and random sampling, and \textbf{FSL} (Fully-Supervised Learning), which serves as the upper bound.
Since the aforementioned TTA benchmarks are released with ResNet-based implementations, their default hyperparameter settings are not directly applicable to transformers, highlighting their sensitivity to hyperparameters across different model architectures. We therefore reimplement them on Swin-T with appropriate hyperparameter choice. Moreover, we exclude RoTTA for Swin-T, as its reliance on robust batch normalization is incompatible with transformer architectures.

\textbf{Baseline Comparison.}
Table \ref{table_pre_trained} summarizes the adaptation performance on the Boston Seaport scenario using both Swin-T and ResNet-50 models. As observed, the proposed \texttt{MoCaFe}/\texttt{p-MoCaFe} framework consistently outperforms other benchmarks significantly.  
\rev{Our method demonstrates particularly significant improvements on Swin-T. We attribute this to Swin-T's strong representation and adaptation capabilities \cite{kim2022broad}, but these capabilities can only be exploited when the supervisory signal is sufficiently reliable. Conventional methods construct adaptation signals from individual-frame confidence, entropy, or neighborhood predictions. When these frame-level signals are corrupted, they may reinforce incorrect predictions. Instead, \texttt{MoCaFe} fuses multiple views of the same object via an information filter to construct a more reliable sequence-level pseudo-label. Moreover, although some baselines employ their own weighting or sample-selection mechanisms, none leverage cross-view inverse-variance fusion and the discrepancy-driven sampling strategy.} 
Besides, \texttt{MoCaFe} achieves performance comparable to the \texttt{p-MoCaFe} framework and surpasses all other baselines across both models. This phenomenon indicates that our approach can achieve superior performance with the \texttt{MoCaFe} scheme without requiring a public dataset.

In addition, we have included \textbf{per-class accuracy} to further demonstrate the effectiveness of our framework across diverse object categories. As shown in Table \ref{table_per_class}, our \texttt{p-MoCaFe} framework substantially outperforms pre-trained models in terms of per-class accuracy, for ResNet-50 with 9 classes improved and only 1 class degraded and for Swin-T with all 10 classes improved. This is because \texttt{p-MoCaFe} is designed to mitigate the impact of noisy pseudo labels: Eq. (\ref{eq:weights}) shows that when the probability of pseudo-label corruption is high, and the distortion is large, a smaller weighting factor $w_f(x)$ shall be assigned to the sample. The simulation results demonstrate that our framework substantially outperforms baselines in combating per-class biases.

\subsection{Impact of Various Key Factors}

\noindent \textbf{Effect of Pre-trained Model Accuracy.}
In Table \ref{table_pre_trained}, we also observe that adaptation performance improvement/decline is greatly influenced by pre-trained model accuracy.
A lower-accuracy model produces more corrupted pseudo labels, which thereby misguide the adaptation process and even cause severe degradation, for example, with Swin-T (65\% pre-trained accuracy), ``w/o MoCaFe'' drops accuracy by over 20\%.
However, our \texttt{MoCaFe} framework demonstrates consistent and remarkable performance improvement even in such challenging scenarios.

\noindent \textbf{Effect of Sequence Construction.}
\rev{Our framework only requires a weak assumption that the same objects can be associated across frames, which can be achieved by class-agnostic tracking based on off-the-shelf object-tracking techniques. To validate our argument, we provide simulation results in Table \ref{table:tracking} to demonstrate the impact of tracking algorithms. 
The sequences are constructed using the
object tracking algorithm (the average multi-object tracking accuracy (AMOTA) in the Boston Seaport scenario 
is 67.3\%) \cite{qin2024towards}. Results indicate that off-the-shelf tracking algorithms are sufficient for sequence construction, with only negligible degradation ($< 0.5\%$) in performance compared to the ideal, ground-truth-based object associations, e.g., 81.23\% to 80.97\% with ResNet-50 and 90.19\% to 89.81\% with Swin-T, demonstrating the practicality of the proposed framework.}

\noindent \textbf{Effect of the Number of Selected Samples.}
Table \ref{sample-table} in Appendix \ref{app:num_nuscenes} illustrates the test accuracy across varying numbers of selected training samples, $m$. \rev{The results show that our \texttt{MoCaFe} and \texttt{p-MoCaFe} frameworks substantially outperform other benchmarks under all evaluated values of $m$. Notably, even when the training data is reduced to $16,000$, \texttt{p-MoCaFe} maintains a robust accuracy of $89.48\%$ on Swin-T and $80.42\%$ on ResNet-50, demonstrating the data efficiency of our approach.}

\begin{table*}[t]
\centering
\caption{Per-class test accuracy in the Boston Seaport scenario on the nuScenes dataset. Pre-trained accuracy for the Boston scenario is 70\%.
$\color{gray}\rightarrow$ represents that the performance remains the same with the pre-trained model.}
\label{table_per_class}
\resizebox{0.90\textwidth}{!}{
\renewcommand{\arraystretch}{1}
\begin{tabular}{l|cccccccccc}
\toprule
& \multicolumn{10}{c}{\textbf{ResNet-50}} \\
Class &Car	&Pedestrian & Traffic-Cone&	Trailer &	Bus	& Motorcycle & Barrier&	Con-Vehicle	&Bicycle	&Truck\\

Pre-trained &82.42 &77.35 &70.46 &77.22 &62.86 &61.11 &53.09 &49.64 &48.15 & 23.28\\
\midrule
w/o MoCaFe  &87.41$\color{green}\uparrow$	&81.05$\color{green}\uparrow$	&76.95$\color{green}\uparrow$  &75.32$\color{red}\downarrow$	&61.90$\color{red}\downarrow$	&47.22$\color{red}\downarrow$	&52.31$\color{red}\downarrow$	&49.68$\color{green}\uparrow$	&46.91$\color{red}\downarrow$	&18.84$\color{red}\downarrow$\\
AdaCon &  88.38$\color{green}\uparrow$
&85.29$\color{green}\uparrow$
&77.42$\color{green}\uparrow$
&\textbf{81.00}$\color{green}\uparrow$
&68.10$\color{green}\uparrow$
&50.05$\color{red}\downarrow$
&53.09$\color{gray}\rightarrow$ 
&40.53$\color{red}\downarrow$
&37.50$\color{red}\downarrow$
&14.64$\color{red}\downarrow$\\
PLUE  & 88.79$\color{green}\uparrow$
&84.98$\color{green}\uparrow$
&75.55$\color{green}\uparrow$
&78.67$\color{green}\uparrow$
&64.87$\color{green}\uparrow$
&55.89$\color{red}\downarrow$
&54.24$\color{green}\uparrow$
&46.15$\color{red}\downarrow$
&39.71$\color{red}\downarrow$
&15.51$\color{red}\downarrow$\\
RoTTA & 87.41$\color{green}\uparrow$ & 83.78$\color{green}\uparrow$ & 73.12$\color{green}\uparrow$ & 
74.74$\color{red}\downarrow$ & 65.39$\color{green}\uparrow$ & 52.53$\color{red}\downarrow$ & \textbf{58.33}$\color{green}\uparrow$ & 40.53$\color{red}\downarrow$ & 37.62$\color{red}\downarrow$ & 12.87$\color{red}\downarrow$\\
C-SFDA &  87.96$\color{green}\uparrow$ & 84.05$\color{green}\uparrow$ & 78.24$\color{green}\uparrow$ & 72.50$\color{red}\downarrow$ & 62.86$\color{gray}\rightarrow$ & \textbf{62.58}$\color{green}\uparrow$ & 54.73$\color{green}\uparrow$ & 51.16$\color{green}\uparrow$ & 35.67$\color{red}\downarrow$ & 18.50$\color{red}\downarrow$ \\
UPA  & 86.43$\color{green}\uparrow$ & 83.02$\color{green}\uparrow$ & 77.84$\color{green}\uparrow$ & 74.97$\color{red}\downarrow$ & 69.52$\color{green}\uparrow$  & 57.67$\color{red}\downarrow$ & 48.84$\color{red}\downarrow$ & 45.86$\color{red}\downarrow$ & 28.88$\color{red}\downarrow$ & 16.73$\color{red}\downarrow$\\
AnCon  & 88.91$\color{green}\uparrow$ & 86.31$\color{green}\uparrow$ & 81.42$\color{green}\uparrow$ & 78.67$\color{green}\uparrow$ & 68.10$\color{green}\uparrow$ & 51.48$\color{red}\downarrow$ & 54.24$\color{green}\uparrow$ & 47.22$\color{red}\downarrow$ & 
42.54$\color{red}\downarrow$ & 16.73$\color{red}\downarrow$\\
\textbf{MoCaFe} 
&90.34$\color{green}\uparrow$	&88.68$\color{green}\uparrow$	&\textbf{83.85}$\color{green}\uparrow$	&76.48$\color{red}\downarrow$	&71.50$\color{green}\uparrow$	&52.81$\color{red}\downarrow$	&54.31$\color{green}\uparrow$	&50.50$\color{green}\uparrow$	&53.14$\color{green}\uparrow$&	19.17$\color{red}\downarrow$\\
\textbf{p-MoCaFe}         
&\textbf{91.19}$\color{green}\uparrow$
&\textbf{89.77}$\color{green}\uparrow$
&83.78$\color{green}\uparrow$	&77.90$\color{green}\uparrow$	&\textbf{73.38}$\color{green}\uparrow$
&61.18$\color{green}\uparrow$
&56.85$\color{green}\uparrow$
&\textbf{56.21}$\color{green}\uparrow$	&\textbf{54.42}$\color{green}\uparrow$
&\textbf{20.71}$\color{red}\downarrow$\\

\midrule
& \multicolumn{10}{c}{\textbf{Swin-T}} \\
Class &Car	&Pedestrian & Traffic-Cone&	Trailer &	Bus	& Motorcycle & Barrier&	Con-Vehicle	&Bicycle	&Truck\\

Pre-trained &85.45 & 	73.42 & 	74.68 & 	72.78 & 	60.95 & 	65.56 & 	52.25 & 	51.76 & 	49.38 & 	24.37 \\
\midrule
w/o MoCaFe   &87.41$\color{green}\uparrow$ & 	78.24$\color{green}\uparrow$ & 	76.05$\color{green}\uparrow$ & 	76.53$\color{green}\uparrow$ & 	70.47$\color{green}\uparrow$ & 	75.56$\color{green}\uparrow$ & 	56.32$\color{green}\uparrow$ & 	49.92$\color{red}\downarrow$& 	46.58$\color{red}\downarrow$&  	19.14$\color{red}\downarrow$ \\
AdaCon &  89.72$\color{green}\uparrow$
&80.02$\color{green}\uparrow$
&78.10$\color{green}\uparrow$
&73.90$\color{green}\uparrow$
&66.83$\color{green}\uparrow$
&73.92$\color{green}\uparrow$
&50.37$\color{red}\downarrow$ 
&45.96$\color{red}\downarrow$
&48.65$\color{red}\downarrow$
&15.13$\color{red}\downarrow$\\
PLUE  & 90.08$\color{green}\uparrow$
&78.10$\color{green}\uparrow$
&78.26$\color{green}\uparrow$
&76.53$\color{green}\uparrow$
&64.43$\color{green}\uparrow$
&68.37$\color{green}\uparrow$
&54.18$\color{green}\uparrow$
&47.31$\color{red}\downarrow$
&49.19$\color{red}\downarrow$
&16.73$\color{red}\downarrow$\\
C-SFDA &  90.30$\color{green}\uparrow$ & 81.23$\color{green}\uparrow$ & 77.41$\color{green}\uparrow$ & 74.55$\color{green}\uparrow$ & 65.95$\color{green}\uparrow$ & 66.83$\color{green}\uparrow$ & 55.83$\color{green}\uparrow$ & 45.66$\color{red}\downarrow$ & 36.59$\color{red}\downarrow$ & 18.96$\color{red}\downarrow$ \\
UPA  & 88.90$\color{green}\uparrow$ & 80.20$\color{green}\uparrow$ & 80.61$\color{green}\uparrow$ & 71.65$\color{red}\downarrow$ & 69.48$\color{green}\uparrow$  & 58.23$\color{red}\downarrow$ & 58.59$\color{green}\uparrow$ & 46.46$\color{red}\downarrow$ & 38.90$\color{red}\downarrow$ & 17.27$\color{red}\downarrow$\\
AnCon  & 92.80$\color{green}\uparrow$ & 84.81$\color{green}\uparrow$ & 82.47$\color{green}\uparrow$ & 75.83$\color{green}\uparrow$ & 67.16$\color{green}\uparrow$ & 62.64$\color{red}\downarrow$ & 52.68$\color{green}\uparrow$ & 43.26$\color{red}\downarrow$ & 
39.18$\color{red}\downarrow$ & 17.36$\color{red}\downarrow$\\
\textbf{MoCaFe}  &93.73$\color{green}\uparrow$ & 	\textbf{91.74}$\color{green}\uparrow$ & 	94.00$\color{green}\uparrow$ & 	90.75$\color{green}\uparrow$ & 	\textbf{90.69}$\color{green}\uparrow$ & 	87.16$\color{green}\uparrow$ & 	85.42$\color{green}\uparrow$ & 	86.23$\color{green}\uparrow$ & 	\textbf{76.12}$\color{green}\uparrow$ & 	35.60$\color{green}\uparrow$ \\
\textbf{p-MoCaFe}   &\textbf{94.20}$\color{green}\uparrow$ & 	91.67$\color{green}\uparrow$ & 	\textbf{94.03}$\color{green}\uparrow$ & 	\textbf{92.74}$\color{green}\uparrow$ & 	90.13$\color{green}\uparrow$ & 	\textbf{88.31}$\color{green}\uparrow$ & 	\textbf{86.88}$\color{green}\uparrow$ & 	\textbf{87.19}$\color{green}\uparrow$ & 	75.41$\color{green}\uparrow$ & 	\textbf{40.60}$\color{green}\uparrow$ \\ 
\bottomrule
\end{tabular}
}
\end{table*}

\subsection{Ablation Study}
\label{subsec:ablation}
We conduct comprehensive ablation studies to validate our design choices, with detailed results and tables provided in Appendix \ref{app:ablation}. Our principal findings can be summarized as follows. \textbf{Robustness of pseudo-label construction}: In comparison to selecting the highest-confidence prediction from each sequence as the teacher (i.e., the max-confidence teacher), the filtered teacher in \texttt{MoCaFe} demonstrates stronger robustness. Specifically, the filtered teacher clearly surpasses the max-confidence teacher when the pre-trained models are very poor (around 40\% accuracy), a regime in which learning from the max-confidence teacher (which can be confidently wrong) may mislead model adaptation. \textbf{Component efficacy}: Data sampling is an indispensable component that constructs effective teacher-student pairs, while the data weighting scheme in \texttt{p-MoCaFe} provides further performance gains. 
\textbf{Robustness to public datasets:} \texttt{p-MoCaFe} is highly insensitive to domain mismatch and the size of the public dataset used for calibration, exhibiting less than a 0.7\% performance drop even when the public dataset size is reduced to 10\%.
\textbf{Hyperparameter stability:} \texttt{p-MoCaFe} remains remarkably stable across different neighborhood sizes ($k$) in $k$-NN estimation. Note that \texttt{MoCaFe} does not introduce any new hyperparameters, and \texttt{p-MoCaFe} only has one hyperparameter $k$ induced by our design. Thus, the robustness to $k$ reveals that our framework is hyperparameter-insensitive.


\begin{table}[t]
\caption{Performance in the Boston Seaport scenario on the nuScenes dataset with sequences constructed by an off-the-shelf object tracking algorithm.}
\label{table:tracking}
\centering
\resizebox{0.93\textwidth}{!}{
\renewcommand{\arraystretch}{1}
\begin{tabular}{l|ccc|ccc}
\toprule
& \multicolumn{3}{c}{\textbf{ResNet-50}} & \multicolumn{3}{|c}{\textbf{Swin-T}} \\
Pre-trained & 75\% & 70\% & 65\%  & 75\% & 70\% & 65\% \\
\midrule
w/o MoCaFe & 75.53$\color{green}\uparrow$($\pm$ 0.24) & 71.41$\color{green}\uparrow$($\pm$ 0.26) & 61.48$\color{red}\downarrow$($\pm$ 0.39) & 76.10$\color{green}\uparrow$($\pm$ 0.25) & 71.96$\color{green}\uparrow$($\pm$ 0.27) & 39.51$\color{red}\downarrow$($\pm$ 0.39)\\
AdaCon & 76.45$\color{green}\uparrow$($\pm$ 0.14) & 72.72$\color{green}\uparrow$($\pm$ 0.18) & 65.46$\color{green}\uparrow$($\pm$ 0.19) & 76.23$\color{green}\uparrow$($\pm$ 0.26) & 72.16$\color{green}\uparrow$($\pm$ 0.22) & 60.39$\color{red}\downarrow$($\pm$ 0.54)  \\
PLUE & 77.16$\color{green}\uparrow$($\pm$ 0.16) & 73.04$\color{green}\uparrow$($\pm$ 0.25) & 66.80$\color{green}\uparrow$($\pm$ 0.26) &  76.88$\color{green}\uparrow$($\pm$ 0.25)     & 73.00$\color{green}\uparrow$($\pm$ 0.32)     & 57.58 $\color{red}\downarrow$($\pm$ 0.34)  \\
RoTTA & 77.35$\color{green}\uparrow$($\pm$ 0.24) & 73.85$\color{green}\uparrow$($\pm$ 0.23) & 67.09$\color{green}\uparrow$($\pm$ 0.27) &---&---&--- \\
C-SFDA & 78.03$\color{green}\uparrow$($\pm$ 0.21) & 73.99$\color{green}\uparrow$($\pm$ 0.28) & 67.60$\color{green}\uparrow$($\pm$ 0.31) & 77.11$\color{green}\uparrow$($\pm$ 0.20) & 71.59$\color{green}\uparrow$($\pm$ 0.23) & 65.42$\color{green}\uparrow$($\pm$ 0.27) \\
UPA & 76.15$\color{green}\uparrow$($\pm$ 0.30) & 73.31$\color{green}\uparrow$($\pm$ 0.31) & 56.81$\color{red}\downarrow$($\pm$ 0.39) & 77.44$\color{green}\uparrow$($\pm$ 0.28) & 72.61$\color{green}\uparrow$($\pm$ 0.27) & 67.55$\color{green}\uparrow$($\pm$ 0.23)\\
AnCon & 78.88$\color{green}\uparrow$($\pm$ 0.24) & 75.04$\color{green}\uparrow$($\pm$ 0.33) & 67.93$\color{green}\uparrow$($\pm$ 0.32) & 
79.73$\color{green}\uparrow$($\pm$ 0.22) & 
75.68$\color{green}\uparrow$($\pm$ 0.27) & 69.88$\color{green}\uparrow$($\pm$ 0.35)\\
\textbf{MoCaFe}  & 
\underline{80.50$\color{green}\uparrow$($\pm$ 0.22)} & \underline{76.73$\color{green}\uparrow$($\pm$ 0.25)} & \underline{69.01$\color{green}\uparrow$($\pm$ 0.26)} &    \underline{89.14$\color{green}\uparrow$($\pm$ 0.19)} & \underline{88.87$\color{green}\uparrow$($\pm$ 0.21)} & \underline{88.78$\color{green}\uparrow$($\pm$ 0.27)}\\
\textbf{p-MoCaFe}  & \textbf{80.97$\color{green}\uparrow$($\pm$ 0.19)} & 
\textbf{77.56$\color{green}\uparrow$($\pm$ 0.21)} & \textbf{69.46$\color{green}\uparrow$($\pm$ 0.25)} &  \textbf{89.81$\color{green}\uparrow$($\pm$ 0.26)} & \textbf{89.62$\color{green}\uparrow$($\pm$ 0.19)} & \textbf{89.59$\color{green}\uparrow$($\pm$ 0.25)} \\
FSL & 88.34$\color{green}\uparrow$($\pm$ 0.13) & 87.40$\color{green}\uparrow$($\pm$ 0.20) & 86.51$\color{green}\uparrow$($\pm$ 0.28) & 91.78$\color{green}\uparrow$($\pm$ 0.10) & 91.68$\color{green}\uparrow$($\pm$ 0.16) & 91.55$\color{green}\uparrow$($\pm$ 0.25)\\
\bottomrule
\end{tabular}
}
\end{table}


\rev{\subsection{Extension to Classification in Detection Models.} \label{sec:detection}
While this work primarily focuses on single-label classification as a proof-of-concept for ``mobility helps learning,'' akin to most TTA schemes~\cite{chen2022contrastive, Litrico_2023_CVPR, chen2024uncertainty, joo2024improving,gong2022note}, modern autonomous systems typically implement object detection models to detect and classify \textit{multiple} objects simultaneously. Our \texttt{MoCaFe} framework seamlessly extends to adapt the classification modules of such models (detailed in Appendix \ref{object_detection}). Using Faster R-CNN~\cite{ren2015faster, wu2019detectron2} as an example, we adapt only its classification module to the Boston Seaport scenario. As detailed in Appendix \ref{object_detection}, \texttt{p-MoCaFe} achieves the highest test accuracy across varying pre-trained model capabilities and sample sizes, with both variants outperforming the baseline by a substantial margin (e.g., $\sim9\%$).}


\section{Conclusion} \label{sec:conclusion}
In this paper, we have proposed a general framework for movable agents to adapt object recognition models without manual labeling. Our idea is to exploit the variance in prediction quality for the same object as the agent moves. Since incorrect object predictions can mislead model adaptation, we have developed a \texttt{MoCaFe} framework to judiciously sample training data and assign weightings, and to reduce stochastic logit variance via multi-view filtering. 
We also proposed an optional extension, \texttt{p-MoCaFe}, to mitigate the impact of pseudo-label corruption and have theoretically shown that the framework yields unbiased learning with respect to the loss function. Experimental results have demonstrated that our proposed adaptation method can significantly improve model accuracy in object classification tasks and outperform existing test-time adaptation schemes.

\noindent \textbf{Limitations and Future Work.} This work serves as a proof-of-concept study for ``mobility helps learning'', focusing on object classification. However, it is promising to extend this idea to other vision tasks, such as object detection and segmentation, because agent movement can provide valuable information for model adaptation in these tasks as well. These aspects can be exploited in future work.

\section*{Acknowledgments.}
\rev{This work was supported in part by the JC STEM Lab of Smart City, funded by The Hong Kong Jockey Club Charities Trust under Contract 2023-0108; the Research Grants Council of the Hong Kong SAR, China (Project No. CityU 11216324); the Hong Kong SAR Government under the Global STEM Professorship; and the Research Grants Council of Hong Kong under Grant 27213824, Grant 17207826, and Grant CRS HKU702/24.}


\bibliographystyle{unsrtnat} 
\bibliography{reference}

@String(CVPR  = {IEEE Conf. Comput. Vis. Pattern Recog.})

@String(CVPR  = {CVPR})

@article{kanamori2003active,
  title={Active learning algorithm using the maximum weighted log-likelihood estimator},
  author={Kanamori, Takafumi and Shimodaira, Hidetoshi},
  journal={Journal of statistical planning and inference},
  volume={116},
  number={1},
  pages={149--162},
  year={2003},
  publisher={Elsevier}
}

@article{iliopoulos2022weighted,
  title={Weighted distillation with unlabeled examples},
  author={Iliopoulos, Fotis and Kontonis, Vasilis and Baykal, Cenk and Menghani, Gaurav and Trinh, Khoa and Vee, Erik},
  journal={Advances in Neural Information Processing Systems},
  volume={35},
  pages={7024--7037},
  year={2022}
}

@misc{oord2019representationlearningcontrastivepredictive,
      title={Representation Learning with Contrastive Predictive Coding}, 
      author={Aaron van den Oord and Yazhe Li and Oriol Vinyals},
      year={2019},
      eprint={1807.03748},
      archivePrefix={arXiv},
      primaryClass={cs.LG},
      url={https://arxiv.org/abs/1807.03748}, 
}

@Article{he2019moco,
  author  = {Kaiming He and Haoqi Fan and Yuxin Wu and Saining Xie and Ross Girshick},
  title   = {Momentum Contrast for Unsupervised Visual Representation Learning},
  journal = {arXiv preprint arXiv:1911.05722},
  year    = {2019},
}

@misc{bachman2019learningrepresentationsmaximizingmutual,
      title={Learning Representations by Maximizing Mutual Information Across Views}, 
      author={Philip Bachman and R Devon Hjelm and William Buchwalter},
      year={2019},
      eprint={1906.00910},
      archivePrefix={arXiv},
      primaryClass={cs.LG},
      url={https://arxiv.org/abs/1906.00910}, 
}

@article{cover1967nearest,
  title={Nearest neighbor pattern classification},
  author={Cover, Thomas and Hart, Peter},
  journal={IEEE transactions on information theory},
  volume={13},
  number={1},
  pages={21--27},
  year={1967},
  publisher={IEEE}
}

@misc{hastie2009elements,
  title={The elements of statistical learning: data mining, inference, and prediction},
  author={Hastie, Trevor},
  year={2009},
  publisher={Springer}
}

@inproceedings{caesar2020nuscenes,
  title={nuscenes: A multimodal dataset for autonomous driving},
  author={Caesar, Holger and Bankiti, Varun and Lang, Alex H and Vora, Sourabh and Liong, Venice Erin and Xu, Qiang and Krishnan, Anush and Pan, Yu and Baldan, Giancarlo and Beijbom, Oscar},
  booktitle={Proceedings of the IEEE/CVF conference on computer vision and pattern recognition},
  pages={11621--11631},
  year={2020}
}

@Article{Samuel59,
  author = 	 "A. L. Samuel",
  title = 	 "Some Studies in Machine Learning Using the Game of
		  Checkers",
  journal =	 "IBM Journal of Research and Development",
  year =	 "1959",
  volume =	 "3",
  number =	 "3",
  pages =	 "211--229"
}

@inproceedings{zhu2023pockengine,
  title={Pockengine: Sparse and efficient fine-tuning in a pocket},
  author={Zhu, Ligeng and Hu, Lanxiang and Lin, Ji and Chen, Wei-Ming and Wang, Wei-Chen and Gan, Chuang and Han, Song},
  booktitle={Proceedings of the 56th Annual IEEE/ACM International Symposium on Microarchitecture},
  pages={1381--1394},
  year={2023}
}

@article{recht2018cifar,
  title={Do cifar-10 classifiers generalize to cifar-10?},
  author={Recht, Benjamin and Roelofs, Rebecca and Schmidt, Ludwig and Shankar, Vaishaal},
  journal={arXiv preprint arXiv:1806.00451},
  year={2018}
}

@article{hinton2015distilling,
  title={Distilling the Knowledge in a Neural Network},
  author={Hinton, Geoffrey},
  journal={arXiv preprint arXiv:1503.02531},
  year={2015}
}

@article{kontonis2024slam,
  title={{SLaM}: Student-label mixing for distillation with unlabeled examples},
  author={Kontonis, Vasilis and Iliopoulos, Fotis and Trinh, Khoa and Baykal, Cenk and Menghani, Gaurav and Vee, Erik},
  journal={Advances in Neural Information Processing Systems},
  volume={36},
  year={2024}
}

@inproceedings{liu2022swin,
  title={Swin transformer v2: Scaling up capacity and resolution},
  author={Liu, Ze and Hu, Han and Lin, Yutong and Yao, Zhuliang and Xie, Zhenda and Wei, Yixuan and Ning, Jia and Cao, Yue and Zhang, Zheng and Dong, Li and others},
  booktitle={Proceedings of the IEEE/CVF conference on computer vision and pattern recognition},
  pages={12009--12019},
  year={2022}
}

@inproceedings{chen2022contrastive,
  title={Contrastive test-time adaptation},
  author={Chen, Dian and Wang, Dequan and Darrell, Trevor and Ebrahimi, Sayna},
  booktitle={Proceedings of the IEEE/CVF Conference on Computer Vision and Pattern Recognition},
  pages={295--305},
  year={2022}
}

@inproceedings{sun2020test,
  title={Test-time training with self-supervision for generalization under distribution shifts},
  author={Sun, Yu and Wang, Xiaolong and Liu, Zhuang and Miller, John and Efros, Alexei and Hardt, Moritz},
  booktitle={International conference on machine learning},
  pages={9229--9248},
  year={2020},
  organization={PMLR}
}

@inproceedings{boudiaf2022parameter,
  title={Parameter-free online test-time adaptation},
  author={Boudiaf, Malik and Mueller, Romain and Ben Ayed, Ismail and Bertinetto, Luca},
  booktitle={Proceedings of the IEEE/CVF Conference on Computer Vision and Pattern Recognition},
  pages={8344--8353},
  year={2022}
}

@article{gong2022note,
  title={Note: Robust continual test-time adaptation against temporal correlation},
  author={Gong, Taesik and Jeong, Jongheon and Kim, Taewon and Kim, Yewon and Shin, Jinwoo and Lee, Sung-Ju},
  journal={Advances in Neural Information Processing Systems},
  volume={35},
  pages={27253--27266},
  year={2022}
}

@inproceedings{agrawal2015learning,
  title={Learning to see by moving},
  author={Agrawal, Pulkit and Carreira, Joao and Malik, Jitendra},
  booktitle={Proceedings of the IEEE international conference on computer vision},
  pages={37--45},
  year={2015}
}

@inproceedings{kundu2020universal,
  title={Universal source-free domain adaptation},
  author={Kundu, Jogendra Nath and Venkat, Naveen and Babu, R Venkatesh and others},
  booktitle={Proceedings of the IEEE/CVF conference on computer vision and pattern recognition},
  pages={4544--4553},
  year={2020}
}

@inproceedings{liang2020we,
  title={Do we really need to access the source data? source hypothesis transfer for unsupervised domain adaptation},
  author={Liang, Jian and Hu, Dapeng and Feng, Jiashi},
  booktitle={International conference on machine learning},
  pages={6028--6039},
  year={2020},
  organization={PMLR}
}

@article{Geiger2013IJRR,
  author = {Andreas Geiger and Philip Lenz and Christoph Stiller and Raquel Urtasun},
  title = {Vision meets Robotics: The KITTI Dataset},
  journal = {International Journal of Robotics Research (IJRR)},
  year = {2013}
}

@InProceedings{Litrico_2023_CVPR,
    author    = {Litrico, Mattia and Del Bue, Alessio and Morerio, Pietro},
    title     = {Guiding Pseudo-Labels With Uncertainty Estimation for Source-Free Unsupervised Domain Adaptation},
    booktitle = {Proceedings of the IEEE/CVF Conference on Computer Vision and Pattern Recognition (CVPR)},
    month     = {June},
    year      = {2023},
    pages     = {7640-7650}
}

@ARTICLE{ma2025sense4fl,
  author={Ma, Yanan and Hu, Senkang and Fang, Zhengru and Ji, Yun and Deng, Yiqin and Fang, Yuguang},
  journal={IEEE Transactions on Mobile Computing}, 
  title={{Sense4FL}: Vehicular Crowdsensing Enhanced Federated Learning for Object Detection in Autonomous Driving}, 
  year={2026},
  volume={25},
  number={8},
  pages={13004-13018},
  doi={10.1109/TMC.2026.3674333}
}

@article{ma2026learning,
  title={Learning from Distributed Eyes: Leveraging Collaborative Perception for Automated Model Adaptation},
  author={Ma, Yanan and Tao, Yihang and Fang, Zhengru and Fang, Zihan and Deng, Yiqin and Chen, Xianhao and Fang, Yuguang},
  journal={arXiv preprint arXiv:2609.18511},
  year={2026}
}

@inproceedings{koh2021wilds,
  title={Wilds: A benchmark of in-the-wild distribution shifts},
  author={Koh, Pang Wei and Sagawa, Shiori and Marklund, Henrik and Xie, Sang Michael and Zhang, Marvin and Balsubramani, Akshay and Hu, Weihua and Yasunaga, Michihiro and Phillips, Richard Lanas and Gao, Irena and others},
  booktitle={International conference on machine learning},
  pages={5637--5664},
  year={2021},
  organization={PMLR}
}

@article{ren2015faster,
  title={Faster r-cnn: Towards real-time object detection with region proposal networks},
  author={Ren, Shaoqing and He, Kaiming and Girshick, Ross and Sun, Jian},
  journal={Advances in neural information processing systems},
  volume={28},
  year={2015}
}

@misc{wu2019detectron2,
  author =       {Yuxin Wu and Alexander Kirillov and Francisco Massa and
                  Wan-Yen Lo and Ross Girshick},
  title =        {Detectron2},
  howpublished = {\url{https://github.com/facebookresearch/detectron2}},
  year =         {2019}
}

@inproceedings{yuan2023robust,
  title={Robust test-time adaptation in dynamic scenarios},
  author={Yuan, Longhui and Xie, Binhui and Li, Shuang},
  booktitle={Proceedings of the IEEE/CVF Conference on Computer Vision and Pattern Recognition},
  pages={15922--15932},
  year={2023}
}

@inproceedings{karim2023c,
  title={C-sfda: A curriculum learning aided self-training framework for efficient source free domain adaptation},
  author={Karim, Nazmul and Mithun, Niluthpol Chowdhury and Rajvanshi, Abhinav and Chiu, Han-pang and Samarasekera, Supun and Rahnavard, Nazanin},
  booktitle={Proceedings of the IEEE/CVF conference on computer vision and pattern recognition},
  pages={24120--24131},
  year={2023}
}

@article{chen2024uncertainty,
  title={Uncertainty-aware pseudo-label filtering for source-free unsupervised domain adaptation},
  author={Chen, Xi and Yang, Haosen and Zhang, Huicong and Yao, Hongxun and Zhu, Xiatian},
  journal={Neurocomputing},
  volume={575},
  pages={127190},
  year={2024},
  publisher={Elsevier}
}

@article{joo2024improving,
  title={Improving self-training under distribution shifts via anchored confidence with theoretical guarantees},
  author={Joo, Taejong and Klabjan, Diego},
  journal={Advances in Neural Information Processing Systems},
  volume={37},
  pages={15246--15271},
  year={2024}
}

@article{martin2021jrdb,
  title={Jrdb: A dataset and benchmark of egocentric robot visual perception of humans in built environments},
  author={Martin-Martin, Roberto and Patel, Mihir and Rezatofighi, Hamid and Shenoi, Abhijeet and Gwak, JunYoung and Frankel, Eric and Sadeghian, Amir and Savarese, Silvio},
  journal={IEEE transactions on pattern analysis and machine intelligence},
  volume={45},
  number={6},
  pages={6748--6765},
  year={2021},
  publisher={IEEE}
}

@inproceedings{voigtlaender2019mots,
  title={Mots: Multi-object tracking and segmentation},
  author={Voigtlaender, Paul and Krause, Michael and Osep, Aljosa and Luiten, Jonathon and Sekar, Berin Balachandar Gnana and Geiger, Andreas and Leibe, Bastian},
  booktitle={Proceedings of the ieee/cvf conference on computer vision and pattern recognition},
  pages={7942--7951},
  year={2019}
}

@inproceedings{fan2019lasot,
  title={Lasot: A high-quality benchmark for large-scale single object tracking},
  author={Fan, Heng and Lin, Liting and Yang, Fan and Chu, Peng and Deng, Ge and Yu, Sijia and Bai, Hexin and Xu, Yong and Liao, Chunyuan and Ling, Haibin},
  booktitle={Proceedings of the IEEE/CVF conference on computer vision and pattern recognition},
  pages={5374--5383},
  year={2019}
}

@inproceedings{qin2024towards,
  title={Towards generalizable multi-object tracking},
  author={Qin, Zheng and Wang, Le and Zhou, Sanping and Fu, Panpan and Hua, Gang and Tang, Wei},
  booktitle={Proceedings of the IEEE/CVF Conference on Computer Vision and Pattern Recognition},
  pages={18995--19004},
  year={2024}
}

@article{zhang2022memo,
  title={Memo: Test time robustness via adaptation and augmentation},
  author={Zhang, Marvin and Levine, Sergey and Finn, Chelsea},
  journal={Advances in neural information processing systems},
  volume={35},
  pages={38629--38642},
  year={2022}
}

@InProceedings{Park_2019_CVPR,
author = {Park, Wonpyo and Kim, Dongju and Lu, Yan and Cho, Minsu},
title = {Relational Knowledge Distillation},
booktitle = {Proceedings of the IEEE/CVF Conference on Computer Vision and Pattern Recognition (CVPR)},
month = {June},
year = {2019}
}

@article{chuang2020debiased,
  title={Debiased contrastive learning},
  author={Chuang, Ching-Yao and Robinson, Joshua and Lin, Yen-Chen and Torralba, Antonio and Jegelka, Stefanie},
  journal={Advances in neural information processing systems},
  volume={33},
  pages={8765--8775},
  year={2020}
}

@inproceedings{caron2021emerging,
  title={Emerging properties in self-supervised vision transformers},
  author={Caron, Mathilde and Touvron, Hugo and Misra, Ishan and J{\'e}gou, Herv{\'e} and Mairal, Julien and Bojanowski, Piotr and Joulin, Armand},
  booktitle={Proceedings of the IEEE/CVF international conference on computer vision},
  pages={9650--9660},
  year={2021}
}

@inproceedings{zhouimage,
  title={Image BERT Pre-training with Online Tokenizer},
  author={Zhou, Jinghao and Wei, Chen and Wang, Huiyu and Shen, Wei and Xie, Cihang and Yuille, Alan and Kong, Tao},
  booktitle={International Conference on Learning Representations},
  year={2022}
}

@article{oquab2024dinov2,
  title={DINOv2: Learning Robust Visual Features without Supervision},
  author={Oquab, Maxime and Darcet, Timoth{\'e}e and Moutakanni, Th{\'e}o and Vo, Huy and Szafraniec, Marc and Khalidov, Vasil and Fernandez, Pierre and Haziza, Daniel and Massa, Francisco and El-Nouby, Alaaeldin and others},
  journal={Transactions on Machine Learning Research Journal},
  year={2024}
}

@inproceedings{zhou2017unsupervised,
  title={Unsupervised learning of depth and ego-motion from video},
  author={Zhou, Tinghui and Brown, Matthew and Snavely, Noah and Lowe, David G},
  booktitle={Proceedings of the IEEE conference on computer vision and pattern recognition},
  pages={1851--1858},
  year={2017}
}

@article{kendall2017uncertainties,
  title={What uncertainties do we need in bayesian deep learning for computer vision?},
  author={Kendall, Alex and Gal, Yarin},
  journal={Advances in neural information processing systems},
  volume={30},
  year={2017}
}

@inproceedings{kim2022broad,
  title={A broad study of pre-training for domain generalization and adaptation},
  author={Kim, Donghyun and Wang, Kaihong and Sclaroff, Stan and Saenko, Kate},
  booktitle={European conference on computer vision},
  pages={621--638},
  year={2022},
  organization={Springer}
}

\newpage
\appendix

\section{Technical Appendices and Supplementary Material}


\subsection{Proof of Lemma \ref{lemma: variance reduction}}
\label{app: variance reduction}
For simplicity of notation in this theoretical analysis, we drop the object index $i$ and consider a generic tracklet of length $L \ge 2$. Under ideal inverse-variance weighting, the reliability weights are $r_l = 1/\sigma_l^2$. Substituting the observation model from Assumption \ref{assumption: logit model} into the expression for the filtered prediction gives:
\begin{equation}
    \boldsymbol{\mu}_L = \frac{\sum_{l=1}^L r_l (\mathbf{s} + \boldsymbol{\epsilon}_l)}{\sum_{l=1}^L r_l}.
\end{equation}
Since the latent clean semantic state $\mathbf{s}$ is invariant across all frames of the tracked object, it can be factored out:
\begin{equation}
    \boldsymbol{\mu}_L = \mathbf{s} + \frac{\sum_{l=1}^L r_l \boldsymbol{\epsilon}_l}{\sum_{l=1}^L r_l}.
\end{equation}
Because the noise terms $\{\boldsymbol{\epsilon}_l\}_{l=1}^L$ are independent Gaussian random vectors, their affine combination remains Gaussian. Furthermore, since their expected value is given by $\mathbb{E}[\boldsymbol{\epsilon}_l] = \mathbf{b}_l$, the expectation of the filtered prediction evaluates to:
\begin{equation}
    \mathbb{E}[\boldsymbol{\mu}_L] = \mathbf{s} + \frac{\sum_{l=1}^L r_l \mathbf{b}_l}{\sum_{l=1}^L r_l} = \mathbf{s} + \bar{\mathbf{b}},
\end{equation}
where $\bar{\mathbf{b}} = \sum_{l=1}^L \alpha_l \mathbf{b}_l$ is the weighted systematic bias, with normalized weights $\alpha_l = \frac{r_l}{\sum_{\tau=1}^L r_\tau}$. 

For the covariance, the independence of the noise terms dictates that cross-covariance terms vanish. Thus, the covariance matrix of $\boldsymbol{\mu}_L$ is given by:
\begin{equation}
    \text{Cov}(\boldsymbol{\mu}_L) = \text{Cov}\left(\frac{\sum_{l=1}^L r_l \boldsymbol{\epsilon}_l}{\sum_{l=1}^L r_l}\right) = \frac{\sum_{l=1}^L r_l^2 \text{Cov}(\boldsymbol{\epsilon}_l)}{\left(\sum_{l=1}^L r_l\right)^2}.
\end{equation}
Substituting $\text{Cov}(\boldsymbol{\epsilon}_l) = \sigma_l^2 \mathbf{I} = \frac{1}{r_l} \mathbf{I}$, the numerator simplifies:
\begin{equation}
    \text{Cov}(\boldsymbol{\mu}_L) = \frac{\sum_{l=1}^L r_l^2 \left(\frac{1}{r_l} \mathbf{I}\right)}{\left(\sum_{l=1}^L r_l\right)^2} = \frac{\sum_{l=1}^L r_l}{\left(\sum_{l=1}^L r_l\right)^2} \mathbf{I} = \left(\sum_{l=1}^L r_l\right)^{-1} \mathbf{I} = \Sigma_L \mathbf{I},
\end{equation}
where $\Sigma_L = \left(\sum_{l=1}^L r_l\right)^{-1} = \left(\sum_{l=1}^L \frac{1}{\sigma_l^2}\right)^{-1}$. Thus, we have shown that $\boldsymbol{\mu}_L \sim \mathcal{N}(\mathbf{s} + \bar{\mathbf{b}}, \Sigma_L \mathbf{I})$.

Finally, because all observation variances are positive ($\sigma_l^2 > 0$), it trivially holds that for any sequence of length $L \ge 2$:
\begin{equation}
    \sum_{l=1}^L \frac{1}{\sigma_l^2} > \max_{1 \le l \le L} \frac{1}{\sigma_l^2}.
\end{equation}
Taking the reciprocal of both sides reverses the inequality, yielding 
\begin{equation}
    \Sigma_L < \min_{1 \le l \le L} \sigma_l^2,
\end{equation}
which completes the proof.

\newpage
\subsection{Proof of Proposition \ref{proposition: error_bound}}
\label{app:error_bound}

The filtered teacher misclassifies an observation if there exists at least one incorrect class $c \neq c^\star$ that achieves a higher fused score than the ground-truth class $c^\star$. The probability of this error is given by:
\begin{equation}
    \mathbb{P}(y^t \neq c^\star) = \mathbb{P}\left(\exists c \neq c^\star: [\boldsymbol{\mu}_L]_c > [\boldsymbol{\mu}_L]_{c^\star}\right).
\end{equation}
Applying the union bound, we can upper-bound this error probability as the sum of pairwise error probabilities:
\begin{equation}
    \mathbb{P}(y^t \neq c^\star) \leq \sum_{c \neq c^\star} \mathbb{P}\left([\boldsymbol{\mu}_L]_c - [\boldsymbol{\mu}_L]_{c^\star} > 0\right).
\end{equation}

For any incorrect class $c \neq c^\star$, define the score difference as $\Delta_c = [\boldsymbol{\mu}_L]_c - [\boldsymbol{\mu}_L]_{c^\star}$. Recall from Lemma \ref{lemma: variance reduction} that the filtered prediction follows a multivariate Gaussian distribution:
\begin{equation}
    \boldsymbol{\mu}_L \sim \mathcal{N}(\mathbf{s} + \bar{\mathbf{b}}, \Sigma_L \mathbf{I}).
\end{equation}
Since the components of $\boldsymbol{\mu}_L$ are independent given the diagonal covariance matrix, the difference $\Delta_c$ is a linear combination of independent Gaussians and is therefore also normally distributed:
\begin{equation}
    \Delta_c \sim \mathcal{N}\left([\mathbf{s} + \bar{\mathbf{b}}]_c - [\mathbf{s} + \bar{\mathbf{b}}]_{c^\star}, 2\Sigma_L\right).
\end{equation}
Substituting the definition of the biased class margin $\Delta_{b,c} = [\mathbf{s} + \bar{\mathbf{b}}]_{c^\star} - [\mathbf{s} + \bar{\mathbf{b}}]_c$, the distribution of $\Delta_c$ simplifies to:
\begin{equation}
    \Delta_c \sim \mathcal{N}(-\Delta_{b,c}, 2\Sigma_L).
\end{equation}

The probability that this difference is greater than zero can be evaluated using the standard normal cumulative distribution function, $\Phi(\cdot)$:
\begin{equation}
    \mathbb{P}(\Delta_c > 0) = \Phi\left(-\frac{\Delta_{b,c}}{\sqrt{2\Sigma_L}}\right).
\end{equation}
Applying the standard Chernoff bound for the Gaussian tail, $\Phi(-z) \leq \exp(-z^2/2)$ for $z > 0$, and substituting $z = \frac{\Delta_{b,c}}{\sqrt{2\Sigma_L}}$, we obtain:
\begin{equation}
    \Phi\left(-\frac{\Delta_{b,c}}{\sqrt{2\Sigma_L}}\right) \leq \exp\left(-\frac{1}{2}\left(\frac{\Delta_{b,c}}{\sqrt{2\Sigma_L}}\right)^2\right) = \exp\left(-\frac{\Delta_{b,c}^2}{4\Sigma_L}\right).
\end{equation}
Substituting this pairwise probability back into the union bound yields the error bound:
\begin{equation}
    \mathbb{P}(y^t \neq c^\star) \le \sum_{c \neq c^\star} \exp\left(-\frac{\Delta_{b,c}^2}{4\Sigma_L}\right).
\end{equation}

Assuming that $\mathbf{b}_l = \mathbf{0}$ for all $l$, the systematic bias vanishes ($\bar{\mathbf{b}} = \mathbf{0}$). Consequently, the biased class margin $\Delta_{b,c}$ reduces to the clean class margin $\Delta_{0,c} = [\mathbf{s}]_{c^\star} - [\mathbf{s}]_c$. The upper bound thus simplifies to:
\begin{equation}
    \mathbb{P}(y^t \neq c^\star) \le \sum_{c \neq c^\star} \exp\left(-\frac{\Delta_{0,c}^2}{4\Sigma_L}\right).
\end{equation}

Recall from Lemma \ref{lemma: variance reduction} that the posterior variance is lower than the minimum individual observation variance, i.e., $\Sigma_L < \sigma_{best}^2$. Because $f(x) = \exp(-A/x)$ is a monotonically increasing function for $A > 0$ and $x > 0$, substituting $\Sigma_L$ with the larger variance $\sigma_{best}^2$ increases the bound:
\begin{equation}
    \sum_{c \neq c^\star} \exp\left(-\frac{\Delta_{0,c}^2}{4\Sigma_L}\right) < \sum_{c \neq c^\star} \exp\left(-\frac{\Delta_{0,c}^2}{4\sigma_{best}^2}\right).
\end{equation}
This establishes that the theoretical misclassification upper bound of the filtered pseudo-label is lower than that of the single best prediction, completing the proof.

\newpage
\subsection{Algorithm}
\label{app:algorithm}

\begin{algorithm}[h]
   \caption{The \texttt{p-MoCaFe} Framework}
   \label{alg:p_mocafe}
   \LinesNumbered
   \KwIn{Target training dataset $\mathcal{X}$, public dataset $\mathcal{S}_\text{v}$, and pre-trained model $f$}
   \KwOut{Adapted model $f^\star$}
   Execute Line 1 to Line 22 in Algorithm~\ref{alg:mocafe_framework} on $\mathcal{X}$ to obtain pseudo-labels $y^t(x)$, filtered logits $\boldsymbol{\mu}(x)$, and subset $\mathcal{X}_m$\;
   $U \leftarrow \emptyset,\; k \leftarrow \left\lceil \sqrt{|\mathcal{S}_\text{v}|}/2 \right\rceil$\;
   Construct filtered pseudo-labels $y^t(x)$ and logits $\boldsymbol{\mu}(x)$ for sequences in $\mathcal{S}_\text{v}$ via Eq.~\eqref{eq:filtered_teacher}\;
   \For{$(x, y) \in \mathcal{S}_\text{v}$}{
$X \leftarrow
\left(\max_c [\sigma(\boldsymbol{\mu}(x))]_c,\,
      \max_c [\sigma(f(x))]_c\right)$\;
       \eIf{$y^t(x) = y$}{
           $(p, d) \leftarrow (0, 1)$\;
       }{
           $(p, d) \leftarrow \left(1,\, \frac{\ell(y^t(x), f(x))}{\ell(y, f(x))}\right)$\;
       }
       $U \leftarrow U \cup \{(X, (p, d))\}$\;
   }
   \For{$x \in \mathcal{X}_m$}{
$Z \leftarrow
\left(\max_c [\sigma(\boldsymbol{\mu}(x))]_c,\,
      \max_c [\sigma(f(x))]_c\right)$\;
       $(\hat{p}, \hat{d}) \leftarrow k\text{-NN}(U, Z)$ using standard Euclidean distance in $\mathbb{R}^2$\;
       $w_f(x) \leftarrow \min \left\{1,\, \frac{\beta(x)}{1 + \hat{p}(\hat{d} - 1)}\right\}$\;
   }
   Adapt model $f$ over $\mathcal{X}_m$ by minimizing $R^w(f)$ via Eq.~\eqref{eq:weighted_risk}\;
\end{algorithm}



\newpage
\subsection{Studying the MSE of a Fixed Prediction}
\label{proof:MSE}

In this section, we study the Mean-Squared-Error (MSE) of a fixed prediction $f$ for an arbitrary object $x \in \mathcal{X}$.
In what follows, we theoretically demonstrate the case when the weighting scheme could potentially underperform the standard unweighted approach from a bias-variance perspective. The derivations also provide theoretical justification for projecting the weights to the $[0,1]$ interval (recall Line 16 of Algorithm \ref{alg:p_mocafe}).
Formally, for the MSE of standard unweighted approach $\operatorname{MSE}(x)$ and the MSE of our proposed \texttt{MoCaFe} scheme $\operatorname{MSE}^w(x)$, the following equations hold
\begin{equation}
    \begin{aligned}
    \operatorname{MSE}(x) & =\mathbb{E}_{y^t(x) \mid x}\left[\left(\ell\left(y, f(x)\right)-\ell(y^t(x), f(x))\right)^2\right], \\
    \operatorname{MSE}^w(x) & =\mathbb{E}_{y^t(x) \mid x}\left[\left(\ell\left(y, f(x)\right)-w_f(x) \ell(y^t(x), f(x))\right)^2\right].
    \end{aligned}
\end{equation}

We have the following proposition.
\begin{proposition} \label{prop:app_1}
    Let $\ell: \mathbb{R}^C \times \mathbb{R}^C \rightarrow \mathbb{R}_{+}$ be a bounded loss function. Fix $x \in \mathcal{X}$ and a predictor $f: \mathcal{X} \rightarrow \mathbb{R}^C$. We have $\operatorname{MSE}(x)<\operatorname{MSE}^w(x)$ if and only if:

    1. $d_f(x)<1 / 2$;
    
    2. $p(x) \in \left(0, \frac{1-2 \cdot d_f(x)}{\left(1-d_f(x)\right)^2}\right)$.

\end{proposition}

Proof: We have:

$$
\begin{aligned}
\operatorname{MSE}(x) & =\mathbb{E}_{y^t(x) \mid x}\left[\left(\ell\left(y, f(x)\right)-\ell(y^t(x), f(x))\right)^2\right] \\
& =p(x)\left(\ell\left(y, f(x)\right)-\ell\left(f_{\text {false}}(x), f(x)\right)\right)^2 \\
& =p(x) \ell\left(y, f(x)\right)^2\left(1-d_f(x)\right)^2
\end{aligned}
$$

and

$$
\begin{aligned}
\operatorname{MSE}^w(x)= & \mathbb{E}_{y^t(x) \mid x}\left[\left(\ell\left(y, f(x)\right)-w_f(x) \ell(y^t(x), f(x))\right)^2\right] \\
= & (1-p(x)) \ell\left(y, f(x)\right)^2\left(1-w_f(x)\right)^2 \\
& +p(x)\left(\ell\left(y, f(x)\right)-w_f(x) \ell\left(f_{\text {false}}(x), f(x)\right)\right)^2 \\
= & (1-p(x)) \ell\left(y, f(x)\right)^2\left(1-w_f(x)\right)^2 \\
& +p(x) \ell\left(y, f(x)\right)^2\left(1-w_f(x) d_f(x)\right)^2.
\end{aligned}
$$


Proposition \ref{prop:app_1} implies that when it is less likely for the teacher to corrupt the label of data sample $x$, i.e., $p(x)$ is sufficiently small, and the prediction of the student is close enough to the corrupted label (i.e., when $d_f(x)$ is small enough), we can use the unweighted estimator instead of the weighted one from a bias-variance trade-off perspective, as the former has smaller MSE in this case. This observation aligns well with our method as we always project $w_f(x)$ to $[0,1]$ ($w_f(x)>1$ iff $d_f(x)<1$ and $\left.p(x)>0\right)$.

\newpage
\subsection{Proof of Proposition \ref{prop:weighted_risk}}
\label{proof:weighted_risk}
Let $\mathbb{X}$ be an unknown distribution over objects, i.e., $x \sim \mathbb{X}$. Also, each $x \in \mathcal{X}_m$ has a ground-truth label $y \in \mathcal{Y}$. Thus, a clean sample follows $\left(x, y\right) \sim \mathbb{\tilde Q}_{\text{clean}}$. To quantify the noise induced by filtered pseudo-labels, we consider a stochastic teacher that outputs a ``corrupted'' label $f_{\text {false}}(x)$ with probability $p(x)$ and the ground-truth label $y$ with probability $1-p(x)$, where $\mathbb{\tilde Q}_{\text{t}}$ denote the induced distribution over data samples and filtered pseudo labels. Considering the weighting strategy in (\ref{eq:weights}), we have

\begin{equation}
\begin{aligned}
\mathbb{E}\left[R^w(f)\right] & =\mathbb{E}_{\mathcal{X}_m \sim \mathbb{\tilde Q}_{\text{t}}}\left[\frac{1}{m} \sum_{x \in \mathcal{X}_m} w_f\left(x\right) \ell\left(y^t(x), f\left(x\right)\right)\right] \\
& =\mathbb{E}_{x \sim \mathbb{X}}\left[\mathbb{E}_{y^t(x) \mid x}\left[w_f(x) \ell(y^t(x), f(x))\right]\right] \\
& =\mathbb{E}_{x \sim \mathbb{X}}\left[w_f(x) \cdot\left(p(x) \ell\left(f_{\text {false}}(x), f(x)\right)+(1-p(x)) \ell\left(y, f(x)\right)\right)\right] \\
& =\mathbb{E}_{x \sim \mathbb{X}}\left[\frac{p(x) \ell\left(f_{\text {false}}(x), f(x)\right)+(1-p(x)) \ell\left(y, f(x)\right)}{1+p(x) \cdot\left(d_f(x)-1\right)}\right] \\
& =\mathbb{E}_{x \sim \mathbb{X}}\left[\frac{\ell\left(y, f(x)\right)+\ell\left(y, f(x)\right) \cdot p(x) \cdot\left(d_f(x)-1\right)}{1+p(x) \cdot\left(d_f(x)-1\right)}\right]. \\
& =R(f).
\end{aligned}
\end{equation}

\subsection{Discussion on the Estimation of (\ref{eq:weights})}
\label{app:estimation}
We provide some discussions on the estimation of (\ref{eq:weights}) and Algorithm \ref{alg:p_mocafe} for practical implementations.

First, the number $k$ of neighbors for weight estimation is set to $k=\lceil{\sqrt{v}}/{2}\rceil$ in Algorithm \ref{alg:p_mocafe}, where $v$ is the size of the public dataset. This is because choosing $k=\alpha\left(v^{2 /(2+\operatorname{dim})}\right)$, where $\operatorname{dim}$ is the dimension of the underlying metric space ($\operatorname{dim}=2$ in our case), is asymptotically optimal, and $\alpha = 1 / 2$ is a popular choice for the constant used in practice \cite{cover1967nearest}, \cite{hastie2009elements}, we also provide the ablation study in Table \ref{tab:knn}.

Second, notice that \eqref{eq:weights} indicates that the weight of an example can exceed 1 if and only if the corresponding distortion value in \eqref{eq:weights} for that example is less than 1. This situation might arise, for instance, when both the student and teacher have the same (or very similar) inaccurate predictions for a given example. In such cases, the weight value in \eqref{eq:weights} suggests that the risk associated with this example should be higher than the low value implied by the unweighted loss function. However, since ground-truth labels are unavailable during training and only the teacher's inaccurate prediction is accessible, assigning a weight greater than 1 in this scenario could lead to the model overfitting the inaccurate label. To prevent this issue, we project all weights onto the $[0, 1]$ interval. Additionally, Appendix \ref{proof:MSE} provides further justification for this projection by analyzing the benefits of constraining weights for low-distortion examples within $[0, 1]$ range using MSE considerations.

At last, according to \eqref{eq:weights}, the weight of an example is a function of the predictor, which, in the case of neural networks, corresponds to the model parameters. 
This means that ideally we should update our weights assignment every time the parameters get updated during training. However, we empirically observe that even estimating the weight assignment only once during the whole training process is sufficient to achieve our objectives. As a result, our method adds minimal overhead to the standard training process. 

\newpage
\subsection{Proof of Proposition \ref{prop: bounded bias}}
\label{app: bounded bias}

Taking the expectation of the clipped weighted empirical risk over the sampling distribution $Q$ and the pseudo-label noise generation process gives:
\begin{equation}
\mathbb{E}[\hat{R}^{\tilde{w}}(f)] = \mathbb{E}_{x \sim Q}\Big[ \tilde{w}_f(x) \mathbb{E}_{y^t|x}[\ell(y^t(x), f(x))] \Big].
\end{equation}
By the pseudo-label noise model, the inner expectation expands to:
\begin{equation}
\mathbb{E}_{y^t|x}[\ell(y^t(x), f(x))] = (1 - p(x))\ell(y, f(x)) + p(x)\ell(f_{\text{false}}(x), f(x)).
\end{equation}
Substituting the definition of the distortion factor, $\ell(f_{\text{false}}(x), f(x)) = d_f(x)\ell(y, f(x))$, we obtain:
\begin{equation}
\mathbb{E}_{y^t|x}[\ell(y^t(x), f(x))] = \big[1 + p(x)(d_f(x) - 1)\big] \ell(y, f(x)).
\end{equation}

For notational simplicity in the subsequent steps, we define the shorthand $a(x) = p(x)(d_f(x) - 1)$ and its empirical estimate $\hat{a}(x) = \hat{p}(x)(\hat{d}_f(x) - 1)$. Hence, the expected empirical risk simplifies to:
\begin{equation}
\mathbb{E}[\hat{R}^{\tilde{w}}(f)] = \mathbb{E}_{x \sim Q}\Big[ \tilde{w}_f(x) \big[1 + a(x)\big] \ell(y, f(x)) \Big].
\end{equation}

Concurrently, via importance sampling, the clean risk can be rewritten under $Q$ as:
\begin{equation}
R(f) = \mathbb{E}_{x \sim \Pi}[\ell(y, f(x))] = \mathbb{E}_{x \sim Q}[\beta(x) \ell(y, f(x))].
\end{equation}
Since $\hat{w}_f(x) = \frac{\beta(x)}{1+\hat{a}(x)}$, it follows that $\beta(x) = \hat{w}_f(x)\big[1+\hat{a}(x)\big]$. Substituting this into the clean risk formulation yields:
\begin{equation}
R(f) = \mathbb{E}_{x \sim Q}\Big[ \hat{w}_f(x) \big[1 + \hat{a}(x)\big] \ell(y, f(x)) \Big].
\end{equation}
Subtracting the true risk from the expected empirical risk isolates the total deviation:
\begin{equation}
\mathbb{E}[\hat{R}^{\tilde{w}}(f)] - R(f) = \mathbb{E}_{x \sim Q}\Big[ \Big(\tilde{w}_f(x)\big[1+a(x)\big] - \hat{w}_f(x)\big[1+\hat{a}(x)\big]\Big) \ell(y, f(x)) \Big].
\end{equation}

To decouple the $k$-NN estimation error from the structural clipping effect, we rewrite the inner term by strategically adding and subtracting $\tilde{w}_f(x)\big[1+\hat{a}(x)\big]$:
\begin{equation}
\tilde{w}_f(x)\big[1+a(x)\big] - \hat{w}_f(x)\big[1+\hat{a}(x)\big] = \tilde{w}_f(x)\big(a(x) - \hat{a}(x)\big) + \big(\tilde{w}_f(x) - \hat{w}_f(x)\big)\big[1+\hat{a}(x)\big].
\end{equation}

Taking the absolute value of the expectation, applying both Jensen's inequality ($|\mathbb{E}[X]| \le \mathbb{E}[|X|]$) and the triangle inequality ($|A+B| \le |A|+|B|$), and substituting $a(x)$ and $\hat{a}(x)$ back to their original definitions yields the stated upper bound:
\begin{equation}
\begin{aligned}
\big|\mathbb{E}[\hat{R}^{\tilde{w}}(f)] - R(f)\big| &\le \mathbb{E}_{x \sim Q}\Big[ \tilde{w}_f(x) \big|p(x)(d_f(x) - 1) - \hat{p}(x)(\hat{d}_f(x) - 1)\big| \ell(y, f(x)) \Big] \\
&\quad + \mathbb{E}_{x \sim Q}\Big[ \big|\tilde{w}_f(x) - \hat{w}_f(x)\big| \big[1+\hat{p}(x)(\hat{d}_f(x) - 1)\big] \ell(y, f(x)) \Big]. 
\end{aligned}    
\end{equation}

\newpage
\section{Experiments}
\subsection{Ablation Study}
\label{app:ablation}

\paragraph{Robustness of Pseudo-label Construction.} 
\rev{
We provide a comprehensive evaluation of pseudo-label robustness using ResNet-50 in Table \ref{tab:pseudo_label_robustness} and Swin-T in Table \ref{tab:pseudo_label_robustness_swint}. 
In comparison to selecting the highest-confidence prediction from each sequence as the teacher (i.e., the max-confidence teacher), the filtered teacher employed in \texttt{MoCaFe} demonstrates stronger robustness. 
In particular, when the pre-trained model is relatively strong (approximately 75\% accuracy), the filtered teacher attains performance comparable to that of max-confidence selection. However, as model quality severely deteriorates, the filtered teacher clearly surpasses the max-confidence teacher when the pre-trained models are very poor (around 40\%/50\% accuracy), a regime in which learning from the max-confidence teacher (which can be confidently wrong) may mislead model adaptation. In contrast, our information filter-based approach mitigates this corruption by aggregating multiple observations and downweighting unreliable views based on cross-view consensus, offering a clear theoretical and empirical advantage.

Specifically, when using ResNet-50 at 75\% initial accuracy, the filtered teacher yields modest gains over max-confidence selection: 0.74 points in pseudo-label correctness and 0.57 points in final test accuracy. However, under severely degraded initializations (42.4\% and 37.4\%), these final accuracy improvements surge to 2.97 and 3.92 points, respectively.
Notably, these gains are substantially larger when employing Swin-T. Our interpretation is that Swin-T has stronger representation and more powerful target-domain adaptation capacity and can therefore exploit improved pseudo-supervision more effectively. For instance, at 75.0\% pretrained accuracy, the filtered teacher improves pseudo-label correctness by 2.18 points and test accuracy by 0.79 points over the max-confidence baseline. Under degraded initializations of 57.0\% and 51.0\%, the final test accuracy improvements expand significantly to 4.57 and 6.01 points, respectively.
}

\begin{table}[h]
    \centering
    \caption{Robustness of pseudo-label construction. We investigate the pseudo-label construction module in \texttt{MoCaFe} with ResNet-50 under the Boston Seaport scenario. ``PL correctness'' measures the correctness percentage of constructed pseudo-labels.}
    \label{tab:pseudo_label_robustness}
    \renewcommand{\arraystretch}{1.1}
    \setlength{\tabcolsep}{4pt} 
    \begin{tabular}{@{}ll|cc@{}}
        \toprule
        \textbf{Pre-train Acc.} & \textbf{Pseudo-labeling Strategy} & \textbf{PL Correctness (\%)} & \textbf{Test Acc. (\%)} \\
        \midrule
        \multirow{2}{*}{75.0\%} 
        & Max-confidence & 84.21 & 80.24 \\
        & Filtered (proposed) & \textbf{84.95} & \textbf{80.81} \\
        \midrule
        \multirow{2}{*}{42.4\%} 
        & Max-confidence & 42.13 & 41.10 \\
        & Filtered (proposed) & \textbf{44.80} & \textbf{44.07} \\
        \midrule
        \multirow{2}{*}{37.4\%} 
        & Max-confidence & 34.85 & 33.12 \\
        & Filtered (proposed) & \textbf{37.73} & \textbf{37.04} \\
        \bottomrule
    \end{tabular}
\end{table}

\begin{table}[h]
    \centering
    \caption{Robustness of pseudo-label construction. We investigate the pseudo-label construction module in \texttt{MoCaFe} with Swin-T under the Boston Seaport scenario.}
    \label{tab:pseudo_label_robustness_swint}
    \renewcommand{\arraystretch}{1.1}
    \setlength{\tabcolsep}{4pt} 
    \begin{tabular}{@{}ll|cc@{}}
        \toprule
        \textbf{Pre-train Acc.} & \textbf{Pseudo-labeling Strategy} & \textbf{PL Correctness (\%)} & \textbf{Test Acc. (\%)} \\
        \midrule
        \multirow{2}{*}{75.0\%} 
        & Max-confidence & 86.54 & 88.81 \\
        & Filtered (proposed) & \textbf{88.72} & \textbf{89.60} \\
        \midrule
        \multirow{2}{*}{57.0\%} 
        & Max-confidence & 64.68 & 68.23 \\
        & Filtered (proposed) & \textbf{68.35} & \textbf{72.80} \\
        \midrule
        \multirow{2}{*}{51.0\%} 
        & Max-confidence & 58.80 & 61.56 \\
        & Filtered (proposed) & \textbf{63.76} & \textbf{67.57} \\
        \bottomrule
    \end{tabular}
\end{table}


\paragraph{Effect of Key Modules.} 
We evaluate the impact of our two core components in Table~\ref{tab:components}. \texttt{MoCaFe} yields a significant performance gain, which demonstrates the effectiveness of the data sampling strategy. However, since \texttt{MoCaFe} lacks the weighting scheme, it performs slightly worse than \texttt{p-MoCaFe}. Moreover, applying our weighting scheme alone with random sampling achieves the worst performance. This indicates that data sampling, which constructs effective teacher-student pairs, is an indispensable component of both \texttt{MoCaFe} and \texttt{p-MoCaFe}.

\begin{table}[h]
    \centering
    \caption{Ablation study on key components of \texttt{p-MoCaFe} in the Boston Seaport scenario on the nuScenes dataset with best in \textbf{bold}.}
    \label{tab:components}
\resizebox{\textwidth}{!}{
\renewcommand{\arraystretch}{1}
\begin{tabular}{cc|ccc|ccc}
\toprule
\multirow{2}{*}{\begin{tabular}{@{}c@{}}\textbf{Data} \\ \textbf{Sampling}\end{tabular}}  & \multirow{2}{*}{\begin{tabular}{@{}c@{}}\textbf{Data} \\ \textbf{Weighting}\end{tabular}}& \multicolumn{3}{c}{\textbf{ResNet-50}} & \multicolumn{3}{|c}{\textbf{Swin-T}} \\
&& 75\% & 70\% & 65\% & 75\% & 70\% & 65\%  \\
\midrule
{\ding{55}} & {\ding{52}} & 77.24$\color{green}\uparrow$($\pm$  0.20) & 73.66$\color{green}\uparrow$($\pm$  0.24) & 65.48$\color{green}\uparrow$($\pm$  0.29)& 77.80$\color{green}\uparrow$($\pm$  0.24) & 74.82$\color{green}\uparrow$($\pm$  0.26) & 43.98$\color{red}\downarrow$($\pm$  0.33) \\
{\ding{52}} & {\ding{55}} & 80.81$\color{green}\uparrow$($\pm$  0.23)& 77.25$\color{green}\uparrow$($\pm$  0.25) & 69.32$\color{green}\uparrow$($\pm$  0.24)& 89.60$\color{green}\uparrow$($\pm$  0.11) & 89.34$\color{green}\uparrow$($\pm$  0.15) & 89.27$\color{green}\uparrow$($\pm$  0.16)\\
{\ding{52}} & {\ding{52}}   &  \textbf{81.23$\color{green}\uparrow$($\pm$  0.13)}& \textbf{77.93$\color{green}\uparrow$($\pm$  0.16)}
& \textbf{69.74$\color{green}\uparrow$($\pm$  0.25)}&  \textbf{90.19$\color{green}\uparrow$($\pm$  0.17)} & \textbf{90.07$\color{green}\uparrow$($\pm$  0.18)} & \textbf{90.02$\color{green}\uparrow$($\pm$  0.12)} \\
\bottomrule
\end{tabular}
}
\end{table}


\paragraph{Effect of Public (Validation) Dataset for \texttt{p-MoCaFe}.}
In our experiments, we use the source public dataset (Singapore Onenorth in nuScenes) for \texttt{p-MoCaFe}, yet it effectively adapted to \textit{diverse} target domains, e.g., Boston Seaport, KITTI dataset, and the embodied agent scenario which has a large domain gap (results in Tables \ref{table_pre_trained}-\ref{table_per_class} and \ref{sample-table}-\ref{table:robot} in Appendix \ref{app:num_nuscenes}-Appendix \ref{app:Embodied}). 
Since the public dataset is used to characterize the relationship between confidence and correctness rather than domain features,
\texttt{p-MoCaFe} is \textit{non-sensitive to the domain mismatch} between the public dataset and the target environment. We also provide an ablation study on the public dataset in Tables \ref{table:valset} and \ref{table:valsize}. Note that, with different public datasets and the size, \texttt{p-MoCaFe} exhibits small variances ($<$0.4\% across datasets and $<$0.7\% performance drop even at 10\% public dataset size) and remains close to \texttt{MoCaFe}.

\begin{table}[htbp]
    \centering
    \begin{minipage}{0.48\textwidth}
        \centering
        \caption{Ablation study on public datasets in \texttt{p-MoCaFe}, where the target domain is Boston Seaport and the pretrained accuracy is 75\%.} 
        \label{table:valset}
        \resizebox{\linewidth}{!}{
            \renewcommand{\arraystretch}{1}
            \begin{tabular}{l|ccc}
                \toprule
                Validation Set  & Singapore & KITTI & Embodied\\
                \midrule
                ResNet-50 & 81.23$\color{green}\uparrow$($\pm$ 0.13) & 81.16$\color{green}\uparrow$($\pm$ 0.26)  & 80.97$\color{green}\uparrow$($\pm$ 0.25) \\
                Swin-T  & 90.19$\color{green}\uparrow$($\pm$ 0.17)  & 90.06$\color{green}\uparrow$($\pm$ 0.22)  & 89.84$\color{green}\uparrow$($\pm$ 0.24) \\
                \bottomrule
            \end{tabular}
        }
    \end{minipage}\hfill 
    \begin{minipage}{0.48\textwidth}
        \centering
        \caption{Ablation study on public dataset size using \texttt{p-MoCaFe} for Boston Seaport. X\% denotes the percentage of the original public dataset size.}
        \label{table:valsize}
        \resizebox{\linewidth}{!}{
            \renewcommand{\arraystretch}{1}
            \begin{tabular}{l|ccc}
                \toprule
                \% of Size & 100\% & 50\% & 10\% \\
                \midrule
                ResNet-50 &  {81.23$\color{green}\uparrow$($\pm$ 0.13)} & {80.92$\color{green}\uparrow$($\pm$ 0.24)}  & {80.71$\color{green}\uparrow$($\pm$ 0.25)}  \\
                Swin-T & {90.19$\color{green}\uparrow$($\pm$ 0.17)}  & {89.67$\color{green}\uparrow$($\pm$ 0.23)}  & {89.52$\color{green}\uparrow$($\pm$ 0.29)}  \\
                \bottomrule
            \end{tabular}
        }
    \end{minipage}
\end{table}

\paragraph{Effect of Hyperparameter $k$.}
For \texttt{MoCaFe}, there is no other framework-induced hyperparameter to adjust across datasets and models. For \texttt{p-MoCaFe}, we show the influence of the neighborhood size $k$ in the $k$-NN method in Table~\ref{tab:knn}. Our framework maintains very stable performance across different $k$. The phenomenon shows that both \texttt{MoCaFe} and \texttt{p-MoCaFe} are insensitive to hyperparameters. This insensitivity stems from our design principle, which does not rely on modifications to the learning objective.

\paragraph{Effect of Availability of $d$.}
Our algorithm can be implemented regardless of whether 
$d$ is available. When $d_{i,l}$ is available (e.g., from the radar), it is used only to assign observations to mobility intervals and therefore a coarse range estimate is sufficient (e.g., $D_j$ is set as 0-40m, 40-80m, >80m). We do not need a precise object-to-agent distance. If distance is unavailable, our method remains effective, as shown in Table \ref{table:u_d}.

\begin{table}[htbp]
    \centering
    \begin{minipage}{0.48\textwidth}
        \centering
        \caption{Ablation study on $k$ using the proposed \texttt{p-MoCaFe} framework in the Boston Seaport scenario with best in \textbf{bold}.}
        \label{tab:knn}
        \resizebox{\linewidth}{!}{ 
            \renewcommand{\arraystretch}{1}
            \begin{tabular}{c|ccc}
                \toprule
                & \multicolumn{3}{c}{\textbf{ResNet-50}} \\
                Pre-trained & 75\% & 70\% & 65\%   \\
                \midrule
                $k = \lceil{\sqrt{v}}/{4} \rceil$   & {81.15 ($\pm$  0.16)}& {77.83 ($\pm$  0.18)} & 69.64 ($\pm$  0.20) \\
                $k = \lceil{\sqrt{v}}/{2}\rceil$ & \textbf{81.23 ($\pm$  0.13)}& \textbf{77.93 ($\pm$  0.16)} & \textbf{69.74 ($\pm$  0.25)} \\
                $k = \lceil\sqrt{v}\rceil$ & 81.19 ($\pm$  0.19) & 77.88 ($\pm$  0.21) & 69.70 ($\pm$  0.28) \\
                \bottomrule
            \end{tabular}
        }
    \end{minipage}\hfill 
    \begin{minipage}{0.48\textwidth}
        \centering
        \caption{Ablation study on the availability of $d$ using the proposed \texttt{MoCaFe} framework in Boston Seaport.}
        \label{table:u_d}
        \resizebox{\linewidth}{!}{
            \renewcommand{\arraystretch}{1}
            \begin{tabular}{l|cccc}
                \toprule
                 & \multicolumn{2}{c}{\textbf{ResNet-50}} & \multicolumn{2}{c}{\textbf{Swin-T}} \\ 
                Pre-trained & 75\% & 65\% & 75\% & 65\% \\ 
                \midrule
                $u$ only & 80.81 ($\pm$  0.25) & 69.28 ($\pm$  0.21) & 89.62 ($\pm$  0.15) & 89.24 ($\pm$  0.13) \\
                $u$ \& $d$ & 80.81 ($\pm$  0.23) & 69.32 ($\pm$  0.24) & 89.60 ($\pm$  0.11) & 89.27 ($\pm$  0.16) \\ 
                \bottomrule
            \end{tabular}
        }
    \end{minipage}
\end{table}

\newpage
\subsection{Extension to Object Detection Models}\label{object_detection}

\subsubsection{Method}
Beyond single-label classification, let us consider extending the \texttt{MoCaFe} framework to adapt the classification module in object detection models.
For instance, a well-known object detection model, Faster R-CNN, has a dedicated classification network, which can be fine-tuned while keeping other parts frozen (i.e., freezing the backbone and the regression network). The main difference from our previous design is that the model can classify multiple objects within a single frame. To employ \texttt{MoCaFe}, we first construct sequences of image frames by their time stamps, where images recorded closely over time are constructed as a sequence. 
A filtered teacher pseudo-label is then constructed for each individual object track across the sequence. Consequently, we calculate the overall discrepancy score $h(x)$ for an entire image frame as the \textit{summation} of the individual discrepancy scores between each object's prediction and its corresponding teacher output. Essentially, the main distinction from single-label classification lies in this aggregated method used to calculate the discrepancy score. The data weighting scheme in \eqref{eq:weights} can still be applied to yield an unbiased estimator of the clean loss. In this way, we seamlessly adapt the informative data sampling and debiased weighting procedures from Sections \ref{subsec: mocafe} and \ref{subsec:p-MoCaFe} to the classification head of object detection models.

\subsubsection{Experimental Results for Faster R-CNN}
\label{fasterRCNN}
In what follows, we provide experimental results by extending our \texttt{MoCaFe} framework to a well-known object detection model, Faster R-CNN with ResNet-50 as the backbone network. Since the TTA baselines in Section \ref{sec:simulation} focus on single-label classification, which cannot be easily adapted to object detection models, we only evaluate our approaches, including \texttt{MoCaFe} and \texttt{p-MoCaFe}, while merely comparing ours with existing TTA schemes in previous single-label classification scenarios.

\begin{figure*}[h]
\centering
    \begin{subfigure}{0.32\textwidth} 
        \includegraphics[width=4cm]{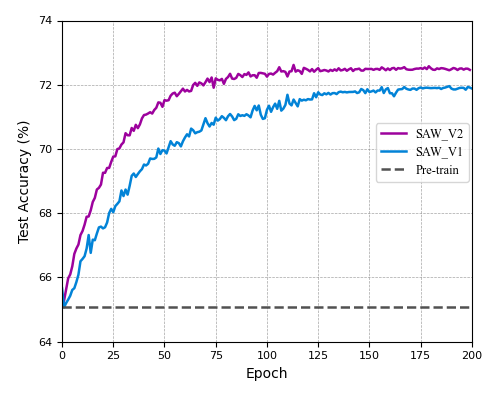}  
        \caption{Pre-trained model accuracy: 65\%}
        \label{fig:acc65swin}
    \end{subfigure}
    \begin{subfigure}{0.32\textwidth} 
        \includegraphics[width=4cm]{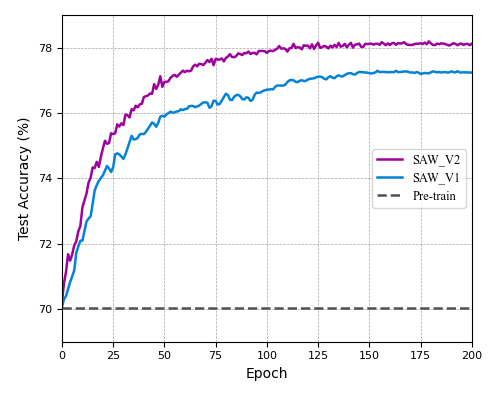}
        \caption{Pre-trained model accuracy: 70\%}
        \label{fig:acc70swin}
   \end{subfigure}
    \begin{subfigure}{0.32\textwidth} 
        \includegraphics[width=4cm]{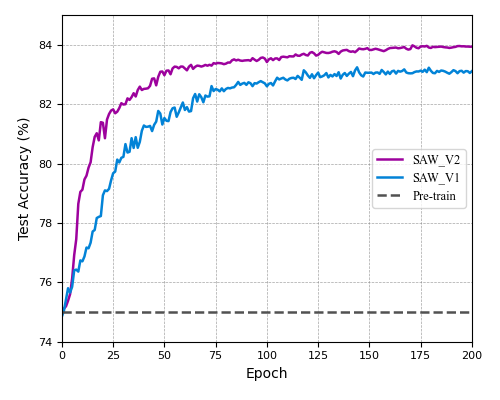}
        \caption{Pre-trained model accuracy: 75\%}
        \label{fig:acc75swin}
   \end{subfigure}\\
  \begin{subfigure}{0.32\textwidth} 
        \includegraphics[width=4cm]{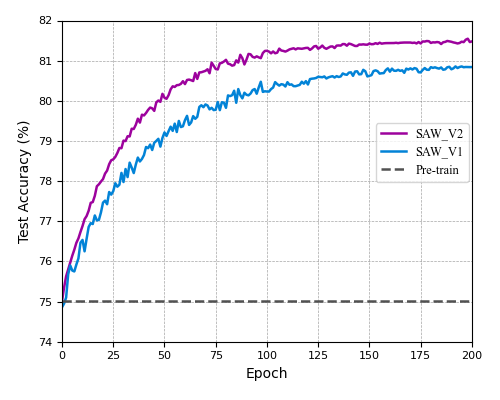}
        \caption{m = 16000}
        \label{fig:m16000swin}
   \end{subfigure}
   \begin{subfigure}{0.32\textwidth} 
        \includegraphics[width=4cm]{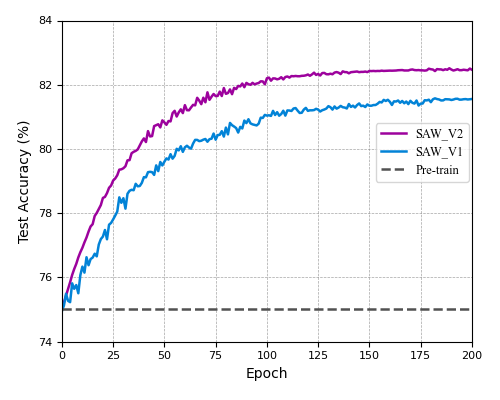}
        \caption{m = 24000}
        \label{fig:m24000swin}
   \end{subfigure}
   \begin{subfigure}{0.32\textwidth} 
        \includegraphics[width=4cm]{figures/faster_acc_75.png}
        \caption{m = 32000}
        \label{fig:m32000swin}
   \end{subfigure}
\caption{The test accuracy versus epoch with the object detection model Faster R-CNN.}
\label{fig:faster_accuracy}
\end{figure*}

As shown in Fig. \ref{fig:faster_accuracy}, \texttt{p-MoCaFe} achieves the best test accuracy across varying settings of pre-trained model accuracy and number of data samples. Moreover, \texttt{p-MoCaFe} and \texttt{MoCaFe} outperform the pre-trained model by a substantial margin (e.g., 9\%). This suggests that our method can be applied to improve the classification modules of object detection models without manual labeling, thereby directly benefiting a wide range of practical applications with movable agents, including autonomous driving, drones, and robots.

\newpage
\subsection{Performance on the Number of Selected Samples on the nuScenes Dataset} \label{app:num_nuscenes}
Table \ref{sample-table} presents the performance of the ResNet-50 and Swin-T architectures on the nuScenes dataset across varying training sample sizes ($m \in \{32,000, 24,000, 16,000\}$). The results demonstrate that our proposed \texttt{MoCaFe} and \texttt{p-MoCaFe} frameworks consistently outperform all other evaluated baselines across all sample-size configurations. Notably, even when the available training data is reduced to 16,000 samples, \texttt{p-MoCaFe} maintains robust adaptation performance, achieving the highest accuracy among all test-time adaptation methods.

\begin{table}[h]
\centering
\caption{Performance in the Boston Seaport scenario on the nuScenes dataset under different numbers of selected training samples $m$. The pre-trained model accuracy for the Boston scenario is 75\%. \textbf{Boldface} and \underline{Underline} highlight the highest and second-highest accuracy, except for the fully-supervised fine-tuning benchmark. 
}
\label{sample-table}
\resizebox{0.96\textwidth}{!}{
\renewcommand{\arraystretch}{1}
\begin{tabular}{l|ccc|ccc}
\toprule
& \multicolumn{3}{c}{\textbf{ResNet-50}} & \multicolumn{3}{|c}{	\textbf{Swin-T}} \\
Num. Samples & $m = 32000$  & $m = 24000$    & $m = 16000$ & $m = 32000$  & $m = 24000$    & $m = 16000$       \\
\midrule 
w/o MoCaFe  &  76.18$\color{green}\uparrow$($\pm$ 0.28) & 75.36$\color{green}\uparrow$($\pm$ 0.15) &  75.31$\color{green}\uparrow$($\pm$ 0.18)    & 76.69$\color{green}\uparrow$($\pm$ 0.26) & 76.06$\color{green}\uparrow$($\pm$ 0.25) & 75.87$\color{green}\uparrow$($\pm$ 0.28) \\
AdaCon &  77.88$\color{green}\uparrow$($\pm$ 0.19) &  76.93$\color{green}\uparrow$($\pm$ 0.23) & 76.27$\color{green}\uparrow$($\pm$ 0.28) & 76.85$\color{green}\uparrow$($\pm$ 0.33)  & 76.24$\color{green}\uparrow$($\pm$ 0.35)  & 75.73$\color{green}\uparrow$($\pm$ 0.37) \\
PLUE &  77.94$\color{green}\uparrow$($\pm$ 0.12) &  76.48$\color{green}\uparrow$($\pm$ 0.23) &  75.84$\color{green}\uparrow$($\pm$ 0.24) & 77.43$\color{green}\uparrow$($\pm$ 0.24)  & 76.95$\color{green}\uparrow$($\pm$ 0.26)  & 76.51$\color{green}\uparrow$($\pm$ 0.25) \\
RoTTA & 78.67$\color{green}\uparrow$($\pm$  0.21) & 77.93$\color{green}\uparrow$($\pm$  0.30) & 77.22$\color{green}\uparrow$($\pm$  0.29) & ----- & ----- & -----\\
C-SFDA & 78.71$\color{green}\uparrow$($\pm$  0.22) & 77.54$\color{green}\uparrow$($\pm$  0.27) & 76.63$\color{green}\uparrow$($\pm$  0.30) & 77.75$\color{green}\uparrow$($\pm$ 0.18)  & 77.36$\color{green}\uparrow$($\pm$ 0.20)  & 
76.97$\color{green}\uparrow$($\pm$ 0.22) \\
UPA & 76.87$\color{green}\uparrow$($\pm$  0.36) & 76.03$\color{green}\uparrow$($\pm$  0.35) & 75.47$\color{green}\uparrow$($\pm$  0.38) & 78.24$\color{green}\uparrow$($\pm$ 0.26)  & 
77.86$\color{green}\uparrow$($\pm$ 0.27) & 77.35$\color{green}\uparrow$($\pm$ 0.26) \\
AnCon & 79.23$\color{green}\uparrow$($\pm$  0.29) & 78.71$\color{green}\uparrow$($\pm$  0.33) & 78.16$\color{green}\uparrow$($\pm$  0.32)& 80.38$\color{green}\uparrow$($\pm$ 0.27)  & 80.02$\color{green}\uparrow$($\pm$ 0.25)  & 
79.66$\color{green}\uparrow$($\pm$ 0.28) \\
\textbf{MoCaFe} & 
\underline{80.81$\color{green}\uparrow$($\pm$ 0.23)} & \underline{80.55$\color{green}\uparrow$($\pm$ 0.18)}  &  \underline{80.01$\color{green}\uparrow$($\pm$ 0.24)}   & \underline{89.60$\color{green}\uparrow$($\pm$ 0.11)}  & \underline{89.22$\color{green}\uparrow$($\pm$ 0.18)}  & \underline{88.94$\color{green}\uparrow$($\pm$ 0.14)} \\
\textbf{p-MoCaFe}  & \textbf{81.23$\color{green}\uparrow$($\pm$ 0.13)} & 
\textbf{80.85$\color{green}\uparrow$($\pm$ 0.18)} & \textbf{80.42$\color{green}\uparrow$($\pm$ 0.26)} & \textbf{90.19$\color{green}\uparrow$($\pm$ 0.17)} & \textbf{89.70$\color{green}\uparrow$($\pm$ 0.14)} & \textbf{89.48$\color{green}\uparrow$($\pm$ 0.25)} \\
FSL  & 88.66$\color{green}\uparrow$($\pm$  0.12)    &87.56$\color{green}\uparrow$($\pm$  0.19)  & 86.83$\color{green}\uparrow$($\pm$  0.28)  & 92.30$\color{green}\uparrow$($\pm$  0.06)   & 91.85$\color{green}\uparrow$($\pm$  0.16) & 91.36$\color{green}\uparrow$($\pm$  0.24)  \\
\bottomrule
\end{tabular}
}
\end{table}

\subsection{Performance on KITTI Dataset} \label{app:kitti}
We also evaluate the performance of our framework by adapting the model to the KITTI dataset, a famous dataset for autonomous driving. 
For adaptation on the KITTI dataset, we follow the same instance sequence construction protocol as in other experiments: objects are extracted from images using ground-truth annotations to form multi-viewpoint sequences. For these, we select $m$ samples to form the adaptation dataset $\mathcal{X}_m$ using the proposed MoCaFe/p-MoCaFe schemes. Similarly, by utilizing perfect sequences, we can isolate the effect of our method from errors introduced by imperfect tracking and demonstrate its empirical potential.

Consistent with the Boston Seaport scenario, we evaluate performance by varying the pre-trained model's accuracy (Table \ref{table_pre_trained_kitti}) and the training sample size $m$ (Table \ref{sample-table_kitti}).
The simulation results demonstrate consistent and substantial performance improvements of the \texttt{MoCaFe} framework compared to pre-trained models across diverse settings. Furthermore, the results also reveal that our \texttt{MoCaFe} and \texttt{p-MoCaFe} substantially outperform all other TTA baselines. It is worth noting that some baselines may achieve even worse performance than the pre-trained model. However, \texttt{MoCaFe} and \texttt{p-MoCaFe} can always improve model performance after adaptation.

\begin{table}[h]
\centering
\caption{Performance on the KITTI dataset under varied pre-trained model accuracy (i.e., 75\%, 70\%, and 65\%). \textbf{Boldface} and \underline{Underline} highlight the highest and second-highest accuracy, except for the fully-supervised fine-tuning benchmark. 
}
\label{table_pre_trained_kitti}
\resizebox{0.96\textwidth}{!}{
\renewcommand{\arraystretch}{1}
\begin{tabular}{l|ccc|ccc}
\toprule
& \multicolumn{3}{c}{\textbf{ResNet-50}} & \multicolumn{3}{|c}{	\textbf{Swin-T}}\\
Pre-train & 75\% & 70\% & 65\% & 75\% & 70\% & 65\%  \\
\midrule
w/o MoCaFe    &76.20$\color{green}\uparrow$($\pm$ 0.24)    &72.14$\color{green}\uparrow$($\pm$ 0.28)    &60.46$\color{red}\downarrow$($\pm$ 0.43) &75.98$\color{green}\uparrow$($\pm$ 0.31)    &72.24$\color{green}\uparrow$($\pm$ 0.28)    &41.42$\color{red}\downarrow$($\pm$ 0.38) \\
AdaCon &  77.03$\color{green}\uparrow$($\pm$ 0.30)    &73.38$\color{green}\uparrow$($\pm$ 0.25)    &60.74$\color{red}\downarrow$($\pm$ 0.32) & 76.32$\color{green}\uparrow$($\pm$ 0.28)  & 
73.51$\color{green}\uparrow$($\pm$ 0.30) & 
59.78$\color{red}\downarrow$($\pm$ 0.43) \\
PLUE & 77.56$\color{green}\uparrow$($\pm$ 0.26)    &73.05$\color{green}\uparrow$($\pm$ 0.31)   &67.32$\color{green}\uparrow$($\pm$ 0.35) & 76.88$\color{green}\uparrow$($\pm$ 0.24)  & 73.40$\color{green}\uparrow$($\pm$ 0.27)  & 
65.32$\color{green}\uparrow$($\pm$ 0.34) \\
RoTTA & 78.16$\color{green}\uparrow$($\pm$ 0.24) & 74.51$\color{green}\uparrow$($\pm$ 0.28) & 67.75$\color{green}\uparrow$($\pm$ 0.39) & ----- & ----- & -----\\
C-SFDA & 78.00$\color{green}\uparrow$($\pm$ 0.20) & 74.98$\color{green}\uparrow$($\pm$ 0.32) & 67.79$\color{green}\uparrow$($\pm$ 0.25) & 77.15$\color{green}\uparrow$($\pm$ 0.23) & 72.87$\color{green}\uparrow$($\pm$ 0.24) & 67.64$\color{green}\uparrow$($\pm$ 0.29) \\
UPA & 76.39$\color{green}\uparrow$($\pm$ 0.37) & 73.44$\color{green}\uparrow$($\pm$ 0.27) & 59.77$\color{red}\downarrow$($\pm$ 0.53) & 76.88$\color{green}\uparrow$($\pm$ 0.19) & 72.57$\color{green}\uparrow$($\pm$ 0.23) & 67.19$\color{green}\uparrow$($\pm$ 0.28)\\
AnCon & 78.46$\color{green}\uparrow$($\pm$ 0.25) & 75.27$\color{green}\uparrow$($\pm$ 0.45) & 68.62$\color{green}\uparrow$($\pm$ 0.37)& 
78.74$\color{green}\uparrow$($\pm$ 0.24)& 74.11$\color{green}\uparrow$($\pm$ 0.27) & 69.04$\color{green}\uparrow$($\pm$ 0.31)\\
\textbf{MoCaFe} 
&\underline{81.04$\color{green}\uparrow$($\pm$ 0.20)}    &\underline{77.95$\color{green}\uparrow$($\pm$ 0.26)}    &\underline{68.80$\color{green}\uparrow$($\pm$ 0.29)} &\underline{84.31$\color{green}\uparrow$($\pm$ 0.15)}    &\underline{83.43$\color{green}\uparrow$($\pm$ 0.18)}   &\underline{83.14$\color{green}\uparrow$($\pm$ 0.26)} \\
\textbf{p-MoCaFe}   
&\textbf{81.37$\color{green}\uparrow$($\pm$ 0.16)}
&\textbf{78.16$\color{green}\uparrow$($\pm$ 0.18)}
&\textbf{69.02$\color{green}\uparrow$($\pm$ 0.29)}
&\textbf{84.79$\color{green}\uparrow$($\pm$ 0.15)}
&\textbf{83.89$\color{green}\uparrow$($\pm$ 0.13)}
&\textbf{83.40$\color{green}\uparrow$($\pm$ 0.18)}\\
FSL &86.98$\color{green}\uparrow$($\pm$  0.15)  &86.12$\color{green}\uparrow$($\pm$  0.27)   &85.25$\color{green}\uparrow$($\pm$  0.19) &88.54$\color{green}\uparrow$($\pm$  0.12)      &87.69$\color{green}\uparrow$($\pm$  0.21)      &86.55$\color{green}\uparrow$($\pm$  0.33)  \\
\bottomrule
\end{tabular}
}
\end{table}

\begin{table}[h]
\caption{Performance on the KITTI dataset under different numbers of selected training samples $m$. The pre-trained model accuracy on the KITTI dataset is 75\%. }
\label{sample-table_kitti}
\centering
\resizebox{0.96\textwidth}{!}{
\renewcommand{\arraystretch}{1}
\begin{tabular}{l|ccc|ccc}
\toprule
& \multicolumn{3}{c}{\textbf{ResNet-50}} & \multicolumn{3}{|c}{\textbf{Swin-T}} \\
Num. Samples & $m = 32000$  & $m = 24000$    & $m = 16000$ & $m = 32000$  & $m = 24000$    & $m = 16000$       \\
\midrule 
w/o MoCaFe  & 76.20$\color{green}\uparrow$($\pm$ 0.24)  & 75.82$\color{green}\uparrow$($\pm$  0.34)  & 75.37$\color{green}\uparrow$($\pm$ 0.31)    & 75.98$\color{green}\uparrow$($\pm$ 0.31)  & 74.71$\color{red}\downarrow$($\pm$ 0.23)  & 74.24$\color{red}\downarrow$($\pm$ 0.24) \\
AdaCon  & 77.03$\color{green}\uparrow$($\pm$ 0.30)  & 76.75$\color{green}\uparrow$($\pm$ 0.21)  & 75.89$\color{green}\uparrow$($\pm$ 0.24)  & 76.32$\color{green}\uparrow$($\pm$ 0.28) & 75.94$\color{green}\uparrow$($\pm$ 0.26) & 75.23$\color{green}\uparrow$($\pm$ 0.34)\\
PLUE &  77.56$\color{green}\uparrow$($\pm$ 0.26)  & 77.11$\color{green}\uparrow$($\pm$ 0.25)  & 76.74$\color{green}\uparrow$($\pm$ 0.20) & 76.88$\color{green}\uparrow$($\pm$ 0.24) & 76.51$\color{green}\uparrow$($\pm$ 0.22) & 75.86$\color{green}\uparrow$($\pm$ 0.28)\\
RoTTA & 78.16$\color{green}\uparrow$($\pm$  0.24) &77.72$\color{green}\uparrow$($\pm$ 0.26)	
&77.30$\color{green}\uparrow$($\pm$ 0.23) & ----- & ----- & -----\\
C-SFDA & 78.00$\color{green}\uparrow$($\pm$  0.20) 
&77.53$\color{green}\uparrow$($\pm$ 0.22)&
77.13$\color{green}\uparrow$($\pm$ 0.27) & 77.15$\color{green}\uparrow$($\pm$ 0.23) & 
76.73$\color{green}\uparrow$($\pm$ 0.25)& 76.20$\color{green}\uparrow$($\pm$ 0.29)\\
UPA & 76.39$\color{green}\uparrow$($\pm$  0.37)
&76.01$\color{green}\uparrow$($\pm$ 0.35)
&75.55$\color{green}\uparrow$($\pm$ 0.38) & 76.88$\color{green}\uparrow$($\pm$ 0.19) & 76.30$\color{green}\uparrow$($\pm$ 0.21) & 75.85$\color{green}\uparrow$($\pm$ 0.27)\\
AnCon & 78.46$\color{green}\uparrow$($\pm$  0.25) 
&78.05$\color{green}\uparrow$($\pm$ 0.23)	
&77.68$\color{green}\uparrow$($\pm$ 0.26)	& 
78.74$\color{green}\uparrow$($\pm$ 0.24) & 78.23$\color{green}\uparrow$($\pm$ 0.19) & 77.88$\color{green}\uparrow$($\pm$ 0.21)\\
\textbf{MoCaFe}  & 
\underline{81.04$\color{green}\uparrow$($\pm$ 0.20)}  & \underline{80.83$\color{green}\uparrow$($\pm$ 0.18)}  & \underline{80.18$\color{green}\uparrow$($\pm$ 0.30)}  & \underline{84.31$\color{green}\uparrow$($\pm$ 0.15)}  & \underline{84.01$\color{green}\uparrow$($\pm$ 0.17)}  & \underline{83.50$\color{green}\uparrow$($\pm$ 0.21)} \\
\textbf{p-MoCaFe} 
&\textbf{81.37$\color{green}\uparrow$($\pm$ 0.16)} &\textbf{80.96$\color{green}\uparrow$($\pm$ 0.22)}
&\textbf{80.59$\color{green}\uparrow$($\pm$ 0.25)} &\textbf{84.79$\color{green}\uparrow$($\pm$ 0.15)} 
&\textbf{84.40$\color{green}\uparrow$($\pm$ 0.17)}
&\textbf{83.91$\color{green}\uparrow$($\pm$ 0.25)} \\
FSL   &86.98$\color{green}\uparrow$($\pm$  0.15)   & 86.10$\color{green}\uparrow$($\pm$  0.24)  & 85.76$\color{green}\uparrow$($\pm$  0.27) & 88.54$\color{green}\uparrow$($\pm$  0.12)   & 87.76$\color{green}\uparrow$($\pm$  0.18)  & 87.28$\color{green}\uparrow$($\pm$  0.29)  \\
\bottomrule
\end{tabular}
}
\end{table}

\newpage
\subsection{Performance on the Embodied Agent Scenario}
\label{app:Embodied}
To further demonstrate the broad applicability of our methodology, we add experiments in the embodied agent domain by adapting the pre-trained models from the Singapore Onenorth scenario. We select frames from the widely investigated tracking datasets, i.e., JRDB \cite{martin2021jrdb} (collected by robots), MOT \cite{voigtlaender2019mots}, and LaSOT \cite{fan2019lasot} datasets, with classes, including car, pedestrian, robot, drone, basketball, bottle, dog, and bicycle, to form a dataset for robotic scenarios. We selected 800 sequences and extracted over 30k images used for model adaptation and 5k images for testing.
It can be seen that our proposed \texttt{MoCaFe} framework consistently outperforms the non-adaptive model and the other baselines. These new experiments in an entirely new domain have demonstrated the broad applicability of our method.

\begin{table}[h]
\caption{Performance on the embodied agent scenario under varied pre-trained model accuracy for this scenario (i.e., 75\%, 70\%, and 65\%). The public dataset is still the Singapore Onenorth scenario on the nuScenes dataset. 
}
\label{table:robot}
\centering
\resizebox{0.96\textwidth}{!}{
\renewcommand{\arraystretch}{1}
\begin{tabular}{l|ccc|ccc}
\toprule
& \multicolumn{3}{c}{\textbf{ResNet-50}} & \multicolumn{3}{|c}{\textbf{Swin-T}}\\
Pre-train & 75\% & 70\% & 65\% & 75\% & 70\% & 65\%  \\
\midrule 
w/o MoCaFe   
& 76.03$\color{green}\uparrow$($\pm$ 0.28) 
& 71.06$\color{green}\uparrow$($\pm$ 0.29) 
& 60.07$\color{red}\downarrow$($\pm$ 0.34)
& 76.03$\color{green}\uparrow$($\pm$ 0.28) 
& 71.32$\color{green}\uparrow$($\pm$ 0.31) 
& 38.47$\color{red}\downarrow$($\pm$ 0.35) \\
AdaCon  
& 76.85$\color{green}\uparrow$($\pm$ 0.26) 
& 72.63$\color{green}\uparrow$($\pm$ 0.25) 
& 65.63$\color{green}\uparrow$($\pm$ 0.29)
& 76.81$\color{green}\uparrow$($\pm$ 0.22) 
& 72.04$\color{green}\uparrow$($\pm$ 0.24) 
& 58.48$\color{red}\downarrow$($\pm$ 0.27)\\
PLUE  
& 77.01$\color{green}\uparrow$($\pm$ 0.25) 
& 72.88$\color{green}\uparrow$($\pm$ 0.27) 
& 65.75$\color{green}\uparrow$($\pm$ 0.33)
& 76.83$\color{green}\uparrow$($\pm$ 0.19) 
& 72.79$\color{green}\uparrow$($\pm$ 0.21)
& 60.27$\color{red}\downarrow$($\pm$ 0.30)\\
RoTTA
&77.76$\color{green}\uparrow$($\pm$ 0.18) 
&74.00$\color{green}\uparrow$($\pm$ 0.20)
&67.36$\color{green}\uparrow$($\pm$ 0.23)
& ---- & ----& ----\\
C-SFDA 
& 77.93$\color{green}\uparrow$($\pm$ 0.22)  
& 73.66$\color{green}\uparrow$($\pm$ 0.22) 
& 68.53$\color{green}\uparrow$($\pm$ 0.31) &77.12$\color{green}\uparrow$($\pm$ 0.17) & 
73.02$\color{green}\uparrow$($\pm$ 0.26)& 
66.88$\color{green}\uparrow$($\pm$ 0.28)\\
UPA 
& 77.57$\color{green}\uparrow$($\pm$ 0.27) 
& 72.61$\color{green}\uparrow$($\pm$ 0.32) 
& 62.23$\color{red}\downarrow$($\pm$ 0.31) & 76.62$\color{green}\uparrow$($\pm$ 0.30)& 72.70$\color{green}\uparrow$($\pm$ 0.29)& 
66.51$\color{green}\uparrow$($\pm$ 0.33)\\
Ancon 
&78.33$\color{green}\uparrow$($\pm$ 0.21) 
& 74.01$\color{green}\uparrow$($\pm$ 0.27) 
& 67.43$\color{green}\uparrow$($\pm$ 0.30)  & 78.89$\color{green}\uparrow$($\pm$ 0.16)& 
75.05$\color{green}\uparrow$($\pm$ 0.26)& 68.84$\color{green}\uparrow$($\pm$ 0.25)\\
\textbf{MoCaFe}
& \underline{79.73$\color{green}\uparrow$($\pm$ 0.15)}
& \underline{77.41$\color{green}\uparrow$($\pm$ 0.20)} 
& \underline{69.03$\color{green}\uparrow$($\pm$ 0.27)}
& \underline{86.98$\color{green}\uparrow$($\pm$ 0.19)} 
& \underline{86.74$\color{green}\uparrow$($\pm$ 0.24)} 
& \underline{86.70$\color{green}\uparrow$($\pm$ 0.26)} \\
\textbf{p-MoCaFe}   
& \textbf{80.18$\color{green}\uparrow$($\pm$ 0.17)} 
& \textbf{77.69$\color{green}\uparrow$($\pm$ 0.23)}
& \textbf{69.37$\color{green}\uparrow$($\pm$ 0.24)}
& \textbf{87.33$\color{green}\uparrow$($\pm$ 0.17)} 
& \textbf{87.22$\color{green}\uparrow$($\pm$ 0.18)}
& \textbf{87.11$\color{green}\uparrow$($\pm$ 0.23)} \\
FSL   
& 86.14$\color{green}\uparrow$($\pm$ 0.15) 
& 85.77$\color{green}\uparrow$($\pm$ 0.16) 
& 84.94$\color{green}\uparrow$($\pm$ 0.20) 
& 89.46$\color{green}\uparrow$($\pm$ 0.10) 
& 89.26$\color{green}\uparrow$($\pm$ 0.17) 
& 89.21$\color{green}\uparrow$($\pm$ 0.20) \\
\bottomrule
\end{tabular}
}
\end{table}

\end{document}